\documentclass[twoside,11pt]{article}

\usepackage{jmlr2e}

\usepackage{subfigure}
\usepackage{amsmath}
\usepackage{url}
\usepackage{color}
\usepackage{verbatim}

\usepackage{lastpage}

\newtheorem{rmk}{Remark}
\newtheorem{assumption}{Assumption}

\newcommand{\be}{\begin{equation}}
\newcommand{\ee}{\end{equation}}
\newcommand{\qedstar}{\hfill $\star$}
\newcommand{\qed}{\hfill $\blacksquare$}

\newcommand{\calD}{ {\cal D}}

\newcommand{\calX}{{\cal X}}

\newcommand{\prob}{\mathbb P}
\newcommand{\probQ}{\mathbb Q}
\newcommand{\probU}{\mathbb U}

\newcommand{\Real}[1]{ { {\mathbb R}^{#1} } }
\newcommand{\dd}{\mathrm{d}}
\newcommand{\ex}{\mathrm{e}}

\newcommand{\eps}{\epsilon}

\newcommand{\sign}{{\mathrm{sign}}}

\jmlrheading{22}{2021}{1-\pageref{LastPage}}{6/21; Revised
	10/21}{11/21}{21-0641}{Marco C. Campi and Simone Garatti}

\ShortHeadings{A Theory of the Risk for Optimization with Relaxation and its Application to SVM}{M.C. Campi and S. Garatti}

\firstpageno{1}

\begin{document}

\title{A Theory of the Risk for Optimization with Relaxation and its Application to Support Vector Machines}

\author{\name Marco C. Campi \email marco.campi@unibs.it \\
       \addr Department of Information Engineering \\
       University of Brescia \\
       via Branze 38, 25123 Brescia, Italy
       \AND
       \name Simone Garatti \email simone.garatti@polimi.it \\
       \addr Dipartimento di Elettronica, Informazione e Bioingegneria \\
       Politecnico di Milano\\
       piazza L. da Vinci 32, 20133 Milano, Italy}

\editor{Corinna Cortes}

\maketitle

\begin{abstract}%
	In this paper we consider optimization with relaxation, an ample paradigm to make data-driven designs. This approach was previously considered by the same authors of this work in \cite{GarCa2019}, a study that revealed a deep-seated connection between two concepts: {\it risk} (probability of not satisfying a new, out-of-sample, constraint) and {\it complexity} (according to a definition introduced in paper \citealp{GarCa2019}). This connection was shown to have profound implications in applications because it implied that the risk can be estimated from the complexity, a quantity that can be measured from the data without any knowledge of the data-generation mechanism. In the present work we establish new results. First, we expand the scope of \cite{GarCa2019} so as to embrace a more general setup that covers various algorithms in machine learning. Then, we study classical support vector methods -- including SVM (Support Vector Machine), SVR (Support Vector Regression) and SVDD (Support Vector Data Description) --  and derive new results for the  ability of these methods to generalize. All results are valid for any finite size of the data set. When the sample size tends to infinity, we establish the unprecedented result that the risk approaches the ratio between the complexity and the cardinality of the data sample, regardless of the value of the complexity.
\end{abstract}


\begin{keywords}
  optimization, optimization with relaxation, generalization, risk quantification,  support vector machines
\end{keywords}

\section{Introduction} \label{sec:intro}

Various techniques in machine learning -- and more generally in data-driven decision-making -- hinge upon the following two ingredients:
\begin{itemize}
	\item[]
	(i) a cost function $c(x)$, which one would like to make as small as possible;\\
	(ii) constraints $f(x,\delta_i) \leq 0$, where $\delta_i$ are observations.
\end{itemize}
In the process of optimizing the cost $c(x)$, constraints $f(x,\delta_i) \leq 0$ can be accounted for in various ways. A flexible paradigm -- which contains more rigid setups as extreme cases -- is obtained by relaxing the constraints and make them ``soft'' according to the following scheme
\begin{align}
\label{scenario program-relaxed}
\min_{x \in \calX \atop \xi_i \geq 0, i=1,\ldots,N} & \quad c(x) + \rho \sum_{i=1}^N \xi_i \\
\textrm{\rm subject to:} & \quad f(x,\delta_i) \leq \xi_i, \ \ i = 1. \ldots ,N. \nonumber
\end{align}
The interpretation of \eqref{scenario program-relaxed} is that some constraints $f(x,\delta_i) \leq 0$ can be violated for the purpose of improving the cost value, but constraints violation has itself a cost as expressed by the auxiliary optimization variables $\xi_i$: if $\xi_i > 0$, then constraint $f(x,\delta_i) \leq 0$ is relaxed to $f(x,\delta_i) \leq \xi_i$ and this generates the regret $\xi_i$, which adds to the original cost $c(x)$. The parameter $\rho$ is used to set a suitable trade-off between the original cost and the cost generated by the regret for violating constraints.

In machine learning, optimization with constraints relaxation plays a major role in various contexts and we provide below examples taken from the gallery of support vector methods, which is a main focus of attention in the present paper.

\begin{example}[Support Vector Regression - SVR]
	Let $\{\delta_i\}_{i=1}^N = \{(\mathbf{u}_i,y_i)\}_{i=1}^N$ be a training set, where the $\mathbf{u}_i$'s are instances living in a suitable input domain, for example $\Real{n}$, and the $y_i$'s are the corresponding output values in $\Real{}$. For given parameters $\tau, \rho > 0$, one considers the optimization program (see e.g. \citealp{ScBaSmWi:98}):
	\begin{align}
	\label{svr}
	\min_{w, \gamma \geq 0, b \in \mathbb{R} \atop \xi_i \geq 0, i=1,\ldots,N} & \quad  (\gamma + \tau \| w \|^2) + \rho \sum_{i=1}^N \xi_i \\
	\textrm{\rm subject to:} & \quad |y_i - \langle w,\mathbf{u}_i \rangle - b | - \gamma \leq \xi_i, \ \ i = 1, \ldots ,N. \nonumber
	\end{align}
	The cost function in \eqref{svr} minimizes a weighted sum of the size $\gamma$ of the ``tube'' used for prediction and the regularization term $\|w\|^2$, to which penalties $\xi_i$ are added for output measurements $y_i$ that are not in the tube (i.e., their distance from the interpolating function $\langle w,\mathbf{u}_i \rangle + b$ is more than $\gamma$). Upon solving program \eqref{svr}, one finds the solution $(w^\ast,\gamma^\ast,b^\ast,\xi_i^\ast)$, which gives the prediction tube
	\begin{equation}
	\label{opt-tube}
	|y - \langle w^\ast,\mathbf{u} \rangle - b^\ast | \leq \gamma^\ast.
	\end{equation}
	When a new value $\mathbf{\bar{u}}$ is received, the corresponding output $\bar{y}$ is forecast to be in the tube, that is, in the range of values of $y$ that satisfy the relation $|y - \langle w^\ast,\mathbf{\bar{u}} \rangle - b^\ast | \leq \gamma^\ast$ and one incurs a prediction error if $\bar{y}$ happens not to belong to this range. \qedstar \\
\end{example}
\begin{example}[Support Vector Data Description - SVDD]
	This is an example of an un-supervised learning technique. Let $\{\delta_i\}_{i=1}^N = \{\mathbf{p}_i\}_{i=1}^N$ be a set of points in $\Real{n}$. SVDD constructs a sphere whose center $c^\ast$ and radius $\gamma^\ast$ are obtained from program (see e.g. \citealp{TaxDuin2004}):
	\begin{align}
	\label{svdd0}
	\min_{c,\gamma \geq 0 \atop \xi_i \geq 0, i=1,\ldots,N} & \quad  \gamma + \rho \sum_{i=1}^N \xi_i \\
	\textrm{\rm subject to:} & \quad \| \mathbf{p}_i - c \|^2 - \gamma \leq \xi_i, \ \ i = 1, \ldots ,N. \nonumber
	\end{align}
	One next out-of-sample point is predicted to be in the sphere and an error is incurred if this does not happen. \qedstar \\
\end{example}
SVR and SVDD can be cast more generally than in the above examples by referring to kernel approaches able to lift the working domain into a feature space. SVR, SVDD, as well as Support Vector Machines (SVM) are studied in detail in Section \ref{sec:svm}, where we apply our new risk theory to derive tight evaluations for the probability of error of these machines. More generally, problem \eqref{scenario program-relaxed} accommodates methods that arise in numerous contexts in data science where relaxation of the constraints can be used to tone down the importance of anomalous observations (sometimes called \emph{outliers}) that would otherwise generate ill-designed solutions, while in other cases relaxation is even strictly necessary to circumvent infeasibility issues (like in SVM with non-linearly separable data). Our theory here developed applies to all these cases.

As previously mentioned, program \eqref{scenario program-relaxed} furnishes a flexible scheme that allows the designer to explore various prospective solutions obtained as $\rho$ varies between the two extremes $\rho = 0$ (no regret for constraints violation) and $\rho = \infty$ (infinite regret for constraints violation, in which case all constraints are rigidly enforced). In this process of selection, one is aided by quantitative tools that describe the quality of the solutions $x^\ast$. Recalling (i) and (ii), it is natural that the designer is concerned about the achieved cost $c(x^\ast)$ and the ensuing \emph{risk} $V(x^\ast)$, where, for a generic value of the optimization variable $x$, the risk
$$
V(x) = \prob \{\delta: f(x,\delta) > 0 \}
$$
($\prob$ is the probability that governs the generation of $\delta$ values) quantifies the probabilistic level of constraints violation (in SVR, violating a constraint corresponds to providing an interval for $\bar{y}$ that does not include its actual value, while in SVDD it amounts to construct a sphere that does not contain the next point). One key-aspect worth noticing is that $c(x^\ast)$ becomes readily available to the designer after the solution $x^\ast$ to problem \eqref{scenario program-relaxed} has been computed; in contrast, the risk of $x^\ast$ cannot be directly evaluated since its definition involves $\prob$, which is normally not available to the user. Hence, evaluating $V(x^\ast)$ requires to develop solid theoretical results and this is instrumental to boost a general trust in data-driven methods, especially in contexts where data are used in automated designs, and not just as a simple support to decisions. The ultimate goal of this contribution is to put forward a new theory that holds true \emph{distribution-free} and yet it allows for tight and practically useful evaluations of the risk.

\subsection{Previous results this paper builds upon}

In \cite{GarCa2019}, the problem of estimating $V(x^\ast)$ was addressed in a convex setup ($c(x)$ and $f(x,\delta)$ are convex in $x$) by adopting the so-called wait-\&-judge perspective of \cite{CamGa2018}. Specifically, a certificate on $V(x^\ast)$ is obtained from the value taken by an observable quantity $s^\ast$, called \emph{complexity} and defined as the number of $\delta_i$'s for which $f(x^\ast,\delta_i) \geq 0$ (i.e., $s^\ast =$ no. of active constraints $+$ no. of violated constraints). Interestingly, the solution $x^\ast$ can be fully reconstructed from the constraints appearing in the definition of $s^\ast$ and, therefore, $s^\ast$ can be interpreted as the complexity of representation of the solution. More discussion on this point is provided in Section \ref{sec:explicit}, where we also offer a systematic and detailed comparison of the results of the present paper with other approaches in the literature.

As is intuitive, the number of violated constraints alone (which, when divided by the number of scenarios, gives the \emph{empirical risk}) is not a valid indicator of the true risk $V(x^\ast)$ since optimization generates a bias towards larger risks by drifting the solution against the constraints. The main achievement of \cite{GarCa2019} consists in showing that the complexity is instead strictly linked to $V(x^\ast)$ and, as such, it can be used to accurately judge the level of risk. This discovery implies a profound and revealing truth: two solutions with the same empirical risk can have quite different true risks $V(x^\ast)$ depending on hidden mechanisms sitting in the method; nonetheless, it is a universal fact that all these mechanisms are captured by the complexity, which, alone, offers an accessible door to evaluate the risk.

Very importantly, applying this theory requires no model for how observations $\delta_i$ are generated. As a matter of fact, although $\delta_i$ are modeled as random outcomes from a probability distribution $\prob$, the obtained results apply irrespective of $\prob$, and $\prob$ remains undefined throughout the algorithmic and theoretical developments of the method. This is practically important since in many applications assuming that $\prob$ is known to the designer is unrealistic: $\prob$ refers to the ``real world'' and can be a truly complex object in modern data science for which hardly complete a-priori knowledge is available (think e.g. of biological or social systems, or of problems arising in autonomous driving, just to cite but a few examples).

To better frame the above mentioned result, we also indicate that the work \cite{GarCa2019} follows in the wake of the so-called ``scenario approach'', initiated with the seminal paper \cite{CalCam:05} and then continued in a stream of theoretical developments, \cite{CamGa:08,	SchiFaMo:13,MarGouLy:14,ZhangEtal:2015,CarGarCam2015}, with application to fields like control system design, \cite{CalCam:06,	SchiFaFrMo2014,GrammaticoEtal:14,FALSONE2019108537}, system identification, \cite{Welsh_Rojas:09,CaCaGa:09,Welsh_Kong:11,Crespo2014CDC,CreKenGie2015,crespo2016interval,LacerdaCrescpoACC2017,GarCamCare2019}, and learning, \cite{Campi-2010,Campi_Care:13,MarPraLyl:14,CarRamCam2018}.

\subsection{New contributions of this paper}

Building upon the achievements of \cite{GarCa2019}, in this paper we establish new results.
\begin{itemize}
	\item[{\bf (a)}] We consider the important class of support vector methods, which have been developed in machine learning for classification and regression problems. In support vector regression methods, the dichotomy between cost and constraints satisfaction described above corresponds to the dichotomy between having informative regressors or classifiers and their probability of misprediction. One contribution of this paper is to establish all the connections between the general theory of \cite{GarCa2019} and support vector methods, including the nontrivial adaptation of the theory to the specific setups when required. It is then shown how the new theory allows for a more reliable usage of support vector methods, especially in relation to the long-standing problem of tuning hyper-parameters, which is key to obtain good solutions.
	\item[{\bf (b)}] Support vector methods are studied in Section \ref{sec:svm}. For a better understanding of this part, we will first revisit in Section \ref{sec:theory} the theory of \cite{GarCa2019} and we will present it in a broader setup than that of \cite{GarCa2019} by considering convex optimization over generic (possibly infinite dimensional) vector spaces. This is a necessary step since generic vector spaces is the natural setup for support vector methods whenever the so called kernel trick is applied. Exploiting the full power of the theory of \cite{GarCa2019} in a general setup is a second contribution of the present paper.\footnote{Note that working in infinite dimensional spaces rules out the possibility of using results where the complexity is \emph{a-priori} bounded by the dimension of the optimization vector as is done, e.g., in \cite{CamGa:08}.}
	\item[{\bf (c)}] We provide asymptotic characterizations of the risk evaluations in (b) and, as a corollary of our theory, Section \ref{sec:explicit} establishes for the first time that the risk of the solution tends  in great generality to the ratio between the complexity of the solution and the sample size $N$, as $N \to \infty$. While our main thrust in this paper remains that of establishing tight evaluations of the risk that are usable for any finite sample size $N$, we remark that this convergence result is unprecedented and sheds new light on the existence of empirical observables that allow one to obtain estimates that converge to the true risk. This new achievement outdoes known results based on measures of complexity of the class of hypotheses as well as results obtained in the domain of compression schemes.
\end{itemize}

\section{Risk Assessment in Optimization with Constraints Relaxation} \label{sec:theory}

In this section, we revisit and extend the theory of \cite{GarCa2019} for the assessment of $V(x^\ast)$ (Theorem \ref{th:concentration-VS-relaxation}). We start by formally stating three assumptions. The first specifies the mathematical frame of work, and the second requires that $x^\ast$ is well-defined. The third assumption is instead a technical requirement whose implications will be commented upon later.
\begin{assumption}[mathematical setup] \label{asmpt:math_setup}
	$x$ is an element of a vector space $\calX$ (possibly infinite dimensional). $c(x)$ and, for any given $\delta$, $f(x,\delta)$ are convex functionals of $x$. The scenarios $\delta_i$, $i=1,\ldots,N$, form an independent and identically distributed (i.i.d.) random sample from a probability space $(\Delta,\calD,\prob)$, that is, $\delta_1, \ldots, \delta_N$ is an outcome from the probability space $(\Delta^N,\calD^N,\prob^N)$, where $\calD^N = \calD \otimes \cdots \otimes \calD$ and $\prob^N = \prob \times \cdots \times \prob$ are the product $\sigma$-algebra and the product probability measure, respectively. \qedstar
\end{assumption}
\begin{assumption}[existence and uniqueness] \label{asmpt:existence-uniqueness-generalized}
	Consider optimization problems as in \eqref{scenario program-relaxed} where $N$ is substituted with any index $m = 0,1,\ldots$ (i.e., $m$ is any nonnegative integer) and $\delta_i$, $i=1,\ldots,m$, is an i.i.d. sample from $(\Delta,\calD,\prob)$. For every $m$ and for every outcome $(\delta_1,\delta_2,\ldots,\delta_m)$, it is assumed that these optimization problems admit a solution (i.e., the problems are feasible and the infimum is achieved on the feasibility set). If for one of these optimization problems more than one solution exists, one solution is singled out by the application of a convex tie-break rule, which breaks the tie by minimizing an additional convex functional $t_1(x)$, and, possibly, other convex functionals $t_2(x)$, $t_3(x)$, \ldots if the tie still occurs.\footnote{Note that only the tie with respect to $x$ is broken by $t_1(x)$, $t_2(x)$, $t_3(x)$, \ldots. On the other hand, for a given $x^\ast$ the values of $\xi_i$, $i=1,\ldots,m$, remain unambiguously determined at optimum by relation $\xi^\ast_{i} = f(x^\ast,\delta_i)$, so that no tie on $\xi_i$, $i=1,\ldots,m$, can persist after the tie on $x$ is broken.} \qedstar
\end{assumption}

The following is a technical non-accumulation assumption of functionals $f(x,\delta)$.
\begin{assumption}[non-accumulation] \label{asmpt: f concentration}
	For every $x$ in $\calX$, $\prob\{\delta: f(x,\delta) = 0\} = 0$. \qedstar
\end{assumption}
This assumption is linked to the concept of non-degeneracy introduced in Definition 3 of \cite{GarCa2019} and it is often satisfied when $\delta$ itself does not accumulate (e.g., when it has a density). Moreover, this assumption is a reasonable modeling simplification even when $\delta$ is discrete but a fine-grained quantity. Examples and more discussion will be provided in Section \ref{sec:svm} in connection to support vector methods.

The following theorem is a reformulation in the present context of the main result of \cite{GarCa2019}.

\begin{theorem}
	\label{th:concentration-VS-relaxation}
	For a given value in $(0,1)$ of the confidence parameter $\beta$, consider for any $k=0,1,\ldots,N-1$ the polynomial equation in the $t$ variable
	\begin{equation}
	\label{pol_eq-for-eps(k)-relax}
	{N \choose k}t^{N-k} - \frac{\beta}{2N} \sum_{i=k}^{N-1} {i \choose k}t^{i-k} - \frac{\beta}{6N} \sum_{i=N+1}^{4N} {i \choose k}t^{i-k}  = 0,
	\end{equation}
	and, for $k = N$, consider the polynomial equation in the $t$ variable
	\begin{equation}
	\label{pol_eq-for-eps(N)-relax}
	1 - \frac{\beta}{6N} \sum_{i=N+1}^{4N} {i \choose N}t^{i-N}  = 0.
	\end{equation}
	For any $k=0,1,\ldots,N-1$, equation \eqref{pol_eq-for-eps(k)-relax} has exactly two solutions in $[0,+\infty)$, which we denote with $\underline{t}(k)$ and $\overline{t}(k)$ ($\underline{t}(k) \leq \overline{t}(k)$). Instead, equation \eqref{pol_eq-for-eps(N)-relax} has only one solution in $[0,+\infty)$, which we denote with $\overline{t}(N)$, while we define $\underline{t}(N) = 0$. Let $\underline{\eps}(k) := \max \{0,1 - \overline{t}(k)\}$ and $\overline{\eps}(k) := 1 - \underline{t}(k)$, $k=0,1,\ldots,N$. Under Assumptions \ref{asmpt:math_setup}, \ref{asmpt:existence-uniqueness-generalized} and \ref{asmpt: f concentration}, for any $\Delta$ and $\prob$ it holds that
	\begin{equation} \label{eq:concentration-VS-relax}
	\prob^N \{ \underline{\eps}(s^\ast) \leq V(x^\ast) \leq \overline{\eps}(s^\ast) \} \geq 1 - \beta,
	\end{equation}
	where $x^\ast$ is the solution to \eqref{scenario program-relaxed}, possibly after breaking the tie according to Assumption \ref{asmpt:existence-uniqueness-generalized}, and $s^\ast$ is the number of $\delta_i$'s for which $f(x^\ast,\delta_i) \geq 0$. \qedstar
\end{theorem}

\begin{proof}
The proof is easily obtained by noticing that the proof of Theorem 4 in \cite{GarCa2019}, given for the case of optimization over Euclidean spaces, applies \emph{mutatis mutandis} to the present more general setup. Details are simple and left to the reader.
\end{proof}

The main message conveyed by Theorem \ref{th:concentration-VS-relaxation} is that it is possible to construct an interval $[\underline{\eps}(s^\ast),\overline{\eps}(s^\ast)]$ where $V(x^\ast)$ lies with high confidence $1-\beta$, and no information on $\Delta$ and $\prob$ is required in this process of evaluation (distribution-free result). The interval depends on $s^\ast$, which is an observable that can be computed from the data record $\delta_1,\ldots,\delta_N$, and for different values of $s^\ast$ one obtains different ranges for $V(x^\ast)$, showing that $s^\ast$ carries fundamental information for the estimation of $V(x^\ast)$.
\begin{figure}
	\centering
	\includegraphics[width=0.6\columnwidth]{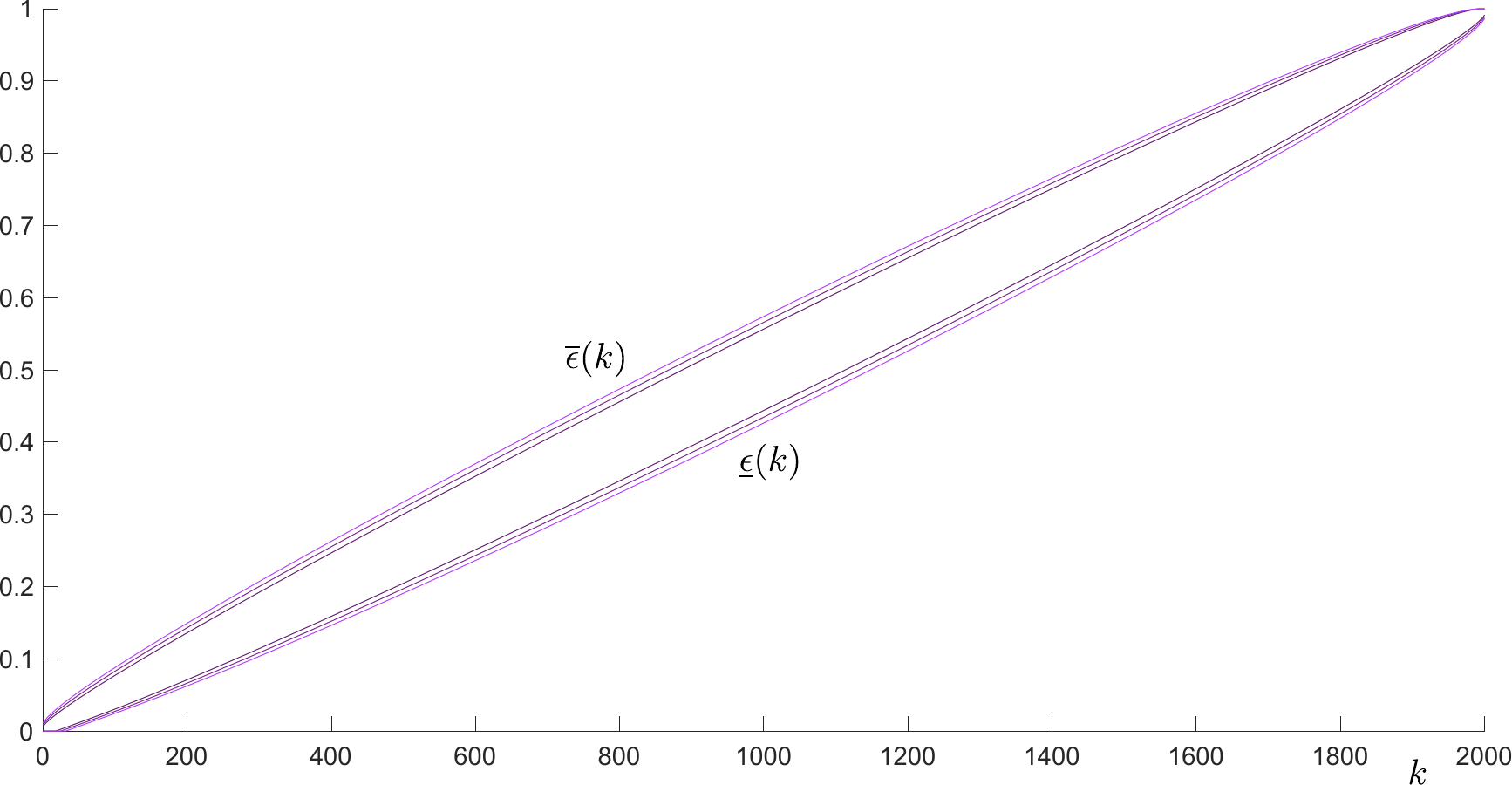}
	\caption{$\underline{\eps}(k)$ and $\overline{\eps}(k)$ for $N=2000$ and $\beta = 10^{-4},10^{-6},10^{-8}$. As $\beta$ decreases, the intervals gently enlarge.}
	\label{fig:epsLU_N_2000_bet_1e-4_-6_-8}
\end{figure}
Figure \ref{fig:epsLU_N_2000_bet_1e-4_-6_-8} depicts $\underline{\eps}(k)$ and $\overline{\eps}(k)$ for $N=2000$ and $\beta = 10^{-4},10^{-6},10^{-8}$, from which we see that small and informative intervals are obtained even for extremely high levels of confidence. Further building on the result in Theorem \ref{th:concentration-VS-relaxation}, in Section \ref{sec:explicit} we shall provide asymptotic evaluations that establish the fact that the risk tends to the ratio between the complexity and the sample size $N$ as $N$ tends to infinity.\footnote{One reason for the tightness of the results in this paper is that these results focus on the risk of the solution rather than being uniform with respect to all potential solutions. This sets an important departure from uniform theories based on the Vapnik-Chervonenkis dimension.}

\begin{figure}
	\centering
	\includegraphics[width=0.7\columnwidth]{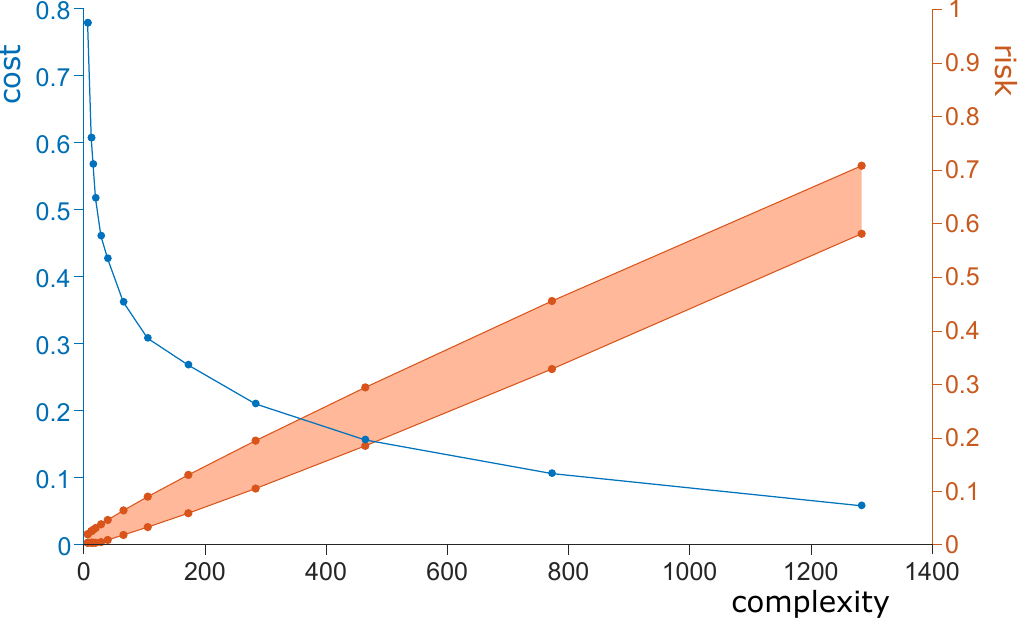}
	\caption{The cost-risk plot. Dots in the picture correspond to the values $k$ of $s^\ast$ that have been observed for a range of selections of the parameter $\rho$. The decreasing function indicates the cost while the intervals show the range for the risk.}
	\label{fig:cost-risk}
\end{figure}

The typical usage of Theorem \ref{th:concentration-VS-relaxation} is as follows. The designer solves \eqref{scenario program-relaxed} repeatedly for various values of $\rho$ and obtains various solutions $x^\ast$ achieving different trade-offs between cost and risk. As $\rho$ varies, the cost is computed, while Theorem \ref{th:concentration-VS-relaxation} allows one to bound the risk based on the observed value of the complexity $s^\ast$. In this way, the designer can generate a cost-risk plot like the one depicted in Figure \ref{fig:cost-risk}, where the cost $c(x^\ast)$ and the interval $[\underline{\eps}(s^\ast),\overline{\eps}(s^\ast)]$ for $V(x^\ast)$ are depicted corresponding to various values of $s^\ast$ (this plot refers to a numerical example presented in Section \ref{sec:numerical_example}).
The user is thus provided with the relevant information to select the solution that achieves the best compromise for the problem at hand. This same reasoning can be carried over to other hyper-parameters besides $\rho$ appearing in the optimization program. As an example, in Section \ref{sec:numerical_example} we shall consider the tuning of the hyper-parameters of a Gaussian kernel.

\begin{rmk}
	A common practice to estimate the risk of the solution consists in dividing the available observations into a training sample $\delta_{i}$, $i=1,\ldots,N_T$, which is used to compute a solution $x^\ast_T$, and a validation sample $\delta_{i}$, $i=N_T\!+\!1,\ldots,N_T\!+\!N_V$, with $N_T\!+\!N_V = N$, by which $V(x^\ast_T)$ is estimated from the ratio 
	\begin{equation}\label{eq:ratio}
	\frac{\mbox{no. of } \delta_i, i=N_T\!+\!1,\ldots,N_T\!+\!N_V, \mbox{ such that } f(x^\ast_T,\delta_i) > 0}{N_V}.
	\end{equation}
	This way of proceeding is justified by the law of large numbers, which ensures that the above ratio tends to $V(x^\ast_T)$ asymptotically, and evaluations of the estimation accuracy can be formulated for any finite $N_V$ as well. However, validation requires sacrificing a portion of the observations to estimate the risk, rather than using them for design purposes. This can be difficult to accept in applications where the data are a valuable and scarce resource. Beyond this point, we feel advisable to offer two more comments that clarify some important theoretical aspects. (i) Using all data to design only moderately reduces the power of the data to achieve the dual effect of generating useful estimates of the risk (refer to our finite sample results and the asymptotic theory in Section \ref{sec:explicit}); (ii) One may be tempted to infer $V(x^\ast)$ (the risk of the solution obtained from all scenarios) from an equation like \eqref{eq:ratio} applied to a solution obtained from a subset of the data (sometimes, it is also suggested to repeatedly apply \eqref{eq:ratio} over, say, $10\%$ leave-out schemes and average the results). However, this way of proceeding is not only invalid from a statistical point of view, it can also generate highly imprecise evaluations when the solution is subject to stochastic variability after leaving out some of the data points.\footnote{This is also true in cross-validation schemes for hyper-parameter selection and is one of the reasons why, besides validation data, one is recommended to save further test data for a final evaluation of the chosen solution.} Instead, Theorem \ref{th:concentration-VS-relaxation} provides a well-principled and statistically valid framework to estimate the risk $V(x^\ast)$, with no waste of information for the design of the solution. \qedstar
\end{rmk}

\begin{rmk}
	For the sake of precision, we feel advisable to point out some extra details in relation to the selection of the hyper-parameter $\rho$. Each single use of Theorem 1 has a probability $\beta$ of providing an incorrect evaluation of the risk. Hence, when the theorem is repeatedly applied to obtain evaluations corresponding to various values of $\rho$, say $p$ values, the probability that the evaluation is wrong in at least one of the values is upper bounded by $p \cdot \beta$. Since we cannot exclude that the user hits a wrong evaluation whenever one exists, the overall procedure is guaranteed with probability $1 - p \cdot \beta$. In other words, using the cost-risk plot to select the value of $\rho$ may result in that the risk is not in the computed interval in at most one case out of $1/ (p \cdot \beta)$. On the other hand, as pointed out in Section \ref{sec:explicit}, enforcing very small values of $\beta$ is ``cheap'' (i.e., it requires moderate numbers of data points) and therefore compensating for the extra probability owing to the increase from $ \beta$ to $p \cdot \beta$ is of little concern for practical purposes. \qedstar
\end{rmk}

\begin{rmk}
It is worth noticing that the result in Theorem \ref{th:concentration-VS-relaxation} has some connection with the theory of conformal prediction of \cite{VovkGammShaf:2005} and \cite{ShaferVovk2008} (see also \citealp{LeiRobinsWasserman2013}, \citealp{Vovk2013}, and \citealp{GyorfiWalk2019} for other contributions). Specifically, it can be shown that the notion of constraint violation for the solution $x^\ast$ implicitly introduces a conformity measure over the scenarios and the risk $V(x^\ast)$ can then be interpreted as a (training sample) conditional coverage. However, rather than using the general tools of conformal prediction that work best for the \emph{mean} of the conditional coverage (also known as unconditional coverage, see \citealp{Vovk2013}), Theorem \ref{th:concentration-VS-relaxation} leverages the specific structure of \eqref{scenario program-relaxed} to provide tight characterizations of the \emph{distribution} of the conditional coverage. This allows one to obtain estimators of $V(x^\ast)$ without resorting to calibration procedures, as done in \cite{Vovk2013}.  \qedstar
\end{rmk}

\begin{rmk}[The price of knowledge]
	Interestingly, reversing the order of the narration, one obtains a new, stimulating, interpretation of Theorem \ref{th:concentration-VS-relaxation}. Theorem \ref{th:concentration-VS-relaxation} claims that small complexity (compared to the sample size $N$) implies reliable models and, viceversa, reliable models requires small complexity. As previously mentioned, the complexity is the size of the sub-sample of observations from which one can re-construct the model. In other words, all other observations become unimportant, and can be discarded, for modeling purposes once this sub-sample is known. A model is our means to describe reality, and it embodies our knowledge about how the portion of reality we are interested in acts and reacts to external stimuli. Hence, Theorem \ref{th:concentration-VS-relaxation} can be interpreted that reliable knowledge within the scheme of \eqref{scenario program-relaxed} can only exist in the presence of an abundance of ineffectual observations. This is what we call the \emph{price of knowledge}.
	
	This reasoning applies broadly to learning schemes that somehow relate to the refutation theory of Popper's philosophy of science, \cite{Popper1962}. Suppose that we construct a model led by a principle of parsimony (for example we build the smallest regression layer limited by straight lines in $\Real{2}$ that contains two-dimensional points whose coordinates are height and weight of observed members of a population). If a new observation agrees with the model (e.g., (height,weight) of a new member falls in the layer), we take it as a confirmation of the model, otherwise, if the new observation does not agree with the model, the model is invalidated (or ``refuted'') and a new model able to accommodate the new observation, besides all observations previously collected, is put in its place. Along this process, if $s^\ast$ remains a small fraction of the total number of observations, the model becomes ``corroborated'' and is expected to survive new invalidation tests as they come along down the stream of observations. In the context we are describing here, the model agrees with all observations (in our formalization, this corresponds to take $\rho = \infty$) and, if we assume that the non-accumulation assumption holds (in e.g. the case of the layer, this follows from requiring that the distribution of points has a density -- see Section \ref{subsection-SVR} -- an assumption that approximately applies in the case of large populations), then we can use Theorem \ref{th:concentration-VS-relaxation} to draw precise and quantitative results supporting the expectation that the model becomes ``corroborated''. Exploring the connections between the mathematical results provided in this paper and broad themes about inductive methods as expressed in the philosophical literature is a goal of great breadth that certainly deserves a much closer attention than that given to it here. \qedstar
\end{rmk}

\subsection{Discussion, asymptotic results, and comparison with the existing literature}
\label{sec:explicit}

Our result, Theorem \ref{th:concentration-VS-relaxation}, can be cast within the frame of work of compression schemes, \cite{FloydWarmuth1995}. In this context, a natural baseline of comparison is \cite{GraepelHerbrichShawe-Taylor2005} in which the so-called luckiness function was introduced to adjust the risk to the value of an observable similarly to our complexity paradigm. Specifically, the result of  \cite{GraepelHerbrichShawe-Taylor2005} (somehow adapted to the framework of our paper) can be summarized as follows. Denote by $c^\ast$ the number of the smallest set of $\delta_i$'s that is sufficient to reconstruct $x^\ast$. As is clear, these  $\delta_i$'s are a subset of  the $\delta_i$'s for which $f(x^\ast,\delta_i) \geq 0$ and, thanks to Assumption \ref{asmpt: f concentration}, they include with probability one all the $\delta_i$'s for which  $f(x^\ast,\delta_i) = 0$. Moreover, let $r^\ast$ be the number of $\delta_i$'s for which $f(x^\ast,\delta_i) > 0$. Note that $s^\ast \leq  c^\ast+r^\ast \leq 2 s^\ast$ with probability one (it can be that $c^\ast + r^\ast$ is strictly greater than $s^\ast$ because a $\delta_i$ can be counted twice as an observation which is needed to reconstruct $x^\ast$ as well as an observation for which $f(x^\ast,\delta_i) > 0$). Then, Theorems 2 and 3 in \cite{GraepelHerbrichShawe-Taylor2005} show that the function
\begin{equation} \label{eq:eps_GHS-T}
\tilde{\eps}(c,r) =
\begin{cases}
\frac{\ln {N \choose c} + \ln N + \ln \frac{1}{\beta}}{N-c}, & c=0,1,\ldots,N, \quad r=0,   \\
\frac{r}{N-c} + \sqrt{\frac{\ln {N \choose c} + 2\ln N + \ln \frac{1}{\beta}}{2(N-c)}}, & c=0,1,\ldots,N, \quad r=1,\ldots,N
\end{cases}
\end{equation}
evaluated in $(c^\ast, r^\ast)$ can be used to upper bound $V(x^\ast)$ according to formula
\begin{equation} \label{eq:V-bound-GHST}
\prob^N \{ V(x^\ast) \leq \tilde{\eps}(c^\ast,r^\ast) \} \geq 1 - \beta.
\end{equation}
Below, we compare $\tilde{\eps}(c^\ast,r^\ast)$ with our $\overline{\eps}(s^\ast)$ appearing in equation \eqref{eq:concentration-VS-relax}. Before doing so, however, we feel advisable to note that having also a lower bound as in our \eqref{eq:concentration-VS-relax} is a guarantee of tightness of our result beyond its comparison with the result provided by \eqref{eq:eps_GHS-T} and \eqref{eq:V-bound-GHST}. Tables \ref{tab:esp_vs_epstilde_1} -- \ref{tab:esp_vs_epstilde_4} show the value of $\tilde{\eps}(c,r)$ against that of $\underline{\eps}(c+r)$ and $\overline{\eps}(c+r)$ for various values of $c$ and $r$.\footnote{Note that this comparison is sharp for situations for which $s^\ast = c^\ast+r^\ast$. It is fair noticing that in many cases $s^\ast < c^\ast+r^\ast$, which introduces a further source of conservatism in the usage of \eqref{eq:V-bound-GHST}.}
{\small
	\begin{table}
		\centering
		\begin{tabular}{|c||c|c|c|c|c|c|}
			\hline
			$r$\textbackslash$c$ & 10 & 20 & 40 & 80 & 160 & 320 \\
			\hline \hline
			0 & 0  -  0.036 & 0 - 0.052 & 0.013 - 0.08 & 0.04 - 0.13 & 0.11 - 0.23 & 0.25 - 0.4 \\
			\hline
			50 & 0.028 - 0.11 & 0.036 - 0.12 & 0.051 - 0.15 & 0.082 - 0.19 & 0.15 - 0.28 & 0.29 - 0.45 \\
			\hline
			100 & 0.066 - 0.17 & 0.074 - 0.18 & 0.09 - 0.2 & 0.12 - 0.25 & 0.19 - 0.34 & 0.34 - 0.51 \\
			\hline
			150 & 0.11 - 0.23 & 0.11 - 0.24 & 0.13 - 0.26 & 0.17 - 0.31 & 0.24 - 0.39 & 0.39 - 0.56 \\
			\hline
			200 & 0.15 - 0.28 & 0.16 - 0.3 &  0.17 - 0.32 & 0.21 - 0.36  & 0.28 - 0.44 & 0.43 - 0.61 \\
			\hline
			250 & 0.19 - 0.34 & 0.20 - 0.35 & 0.22 - 0.37 & 0.25 - 0.41 & 0.33 - 0.49 & 0.48 - 0.65 \\
			\hline
		\end{tabular}
		\caption{Values of $\underline{\eps}(c+r)$ and $\overline{\eps}(c+r)$; $N=1000$, $\beta = 10^{-5}$.}
		\label{tab:esp_vs_epstilde_1}
	\end{table}
	\begin{table}
		\centering
		\begin{tabular}{|c||c|c|c|c|c|c|}
			\hline
			$r$\textbackslash$c$  & 10 & 20 & 40 & 80 & 160 & 320 \\
			\hline \hline
			0 & 0.073 & 0.116 & 0.191 & 0.319 & 0.54 & 0.944 \\
			\hline
			50 & 0.251 & 0.299 & 0.367 & 0.459 & 0.584 & 0.764 \\
			\hline
			100 & 0.301 & 0.351 & 0.419 & 0.513 & 0.643 & 0.837 \\
			\hline
			150 & 0.352 & 0.402 & 0.471 & 0.568 & 0.703 & 0.911 \\
			\hline
			200 & 0.402 & 0.453 & 0.523 & 0.622 & 0.762 & 0.985 \\
			\hline
			250 & 0.453 & 0.504 & 0.575 & 0.676 & 0.822 & 1 \\
			\hline	
		\end{tabular}
		\caption{Values of $\tilde{\eps}(c,r)$; $N=1000$, $\beta = 10^{-5}$.}
		\label{tab:esp_vs_epstilde_2}
	\end{table}
	\begin{table}
		\centering
		\begin{tabular}{|c||c|c|c|c|c|c|}
			\hline
			$r$\textbackslash$c$ & 10 & 20 & 40 & 80 & 160 & 320 \\
			\hline \hline
			0 & 0 - 0.018 &  0 - 0.026 & 0.007 - 0.041 & 0.021 - 0.067 & 0.052 - 0.12 & 0.12 - 0.21 \\
			\hline
			50 & 0.014 - 0.055 & 0.018 - 0.061 & 0.025 - 0.074 & 0.04 - 0.098 & 0.073 - 0.15 & 0.14 - 0.23 \\
			\hline
			100 & 0.033 - 0.086 & 0.036 - 0.092 & 0.044 - 0.10 & 0.06 - 0.13 & 0.09 - 0.17 & 0.16 - 0.26 \\
			\hline
			150 & 0.052 - 0.12 & 0.056 - 0.12 & 0.065 - 0.13 & 0.081 - 0.16 & 0.12 - 0.2 & 0.19 - 0.29 \\
			\hline
			200 & 0.073 - 0.15 & 0.077 - 0.15 & 0.085 - 0.16 & 0.1 - 0.19  & 0.14 - 0.23 & 0.21 - 0.32 \\
			\hline
			250 & 0.094 - 0.17 & 0.098 - 0.18 & 0.11 - 0.19 & 0.12 - 0.21 & 0.16 - 0.26 & 0.23 - 0.34 \\
			\hline
		\end{tabular}
		\caption{Values of $\underline{\eps}(c+r)$ and $\overline{\eps}(c+r)$; $N=2000$, $\beta = 10^{-5}$.}
		\label{tab:esp_vs_epstilde_3}
	\end{table}
	\begin{table}
		\centering
		\begin{tabular}{|c||c|c|c|c|c|c|}
			\hline
			$r$\textbackslash$c$  & 10 & 20 & 40 & 80 & 160 & 320 \\
			\hline \hline
			0 & 0.04 & 0.065 & 0.108 & 0.183 & 0.31 & 0.533 \\
			\hline
			50 & 0.174 &  0.211 & 0.262 & 0.332 & 0.425 & 0.548 \\
			\hline
			100 & 0.199 & 0.236 & 0.288 & 0.358 & 0.452 & 0.578 \\
			\hline
			150 & 0.224 & 0.261 & 0.314 & 0.384 & 0.479  & 0.607 \\
			\hline
			200 & 0.249 & 0.287 & 0.339 & 0.41 &  0.506 & 0.637 \\
			\hline
			250 & 0.274 & 0.312 & 0.365 & 0.436 & 0.533 & 0.667
			\\
			\hline	
		\end{tabular}
		\caption{Values of $\tilde{\eps}(c,r)$; $N=2000$, $\beta = 10^{-5}$.}
		\label{tab:esp_vs_epstilde_4}
	\end{table}
}

We next move to study the asymptotic behavior of our bounds $\overline{\eps}(k)$ and $\underline{\eps}(k)$ as $N \to \infty$.

\begin{theorem}
	\label{th:explicit}
	Functions $\underline{\eps}(k)$ and $\overline{\eps}(k)$ introduced in Theorem \ref{th:concentration-VS-relaxation} are subject to the following bounds:
	\begin{equation} \label{eq:epsU<=}
	\overline{\eps}(k) \leq \frac{k}{N} + C \frac{\sqrt{k} \ln \frac{1}{\beta} + \sqrt{k} \ln k + 1 }{N}
	\end{equation}
	\begin{equation} \label{eq:epsU>=}
	\underline{\eps}(k) \geq \frac{k}{N} - C \frac{\sqrt{k} \ln \frac{1}{\beta} + \sqrt{k} \ln k + 1 }{N}
	\end{equation}
	where $C$ is a suitable constant (independent of $k$, $N$ and $\beta$) and the bounds hold for $1 \leq k \leq N$ and $\beta \in (0,1)$, while, for $k = 0$, we have $\overline{\eps}(0) \leq (\ln(1/\beta) + 1) \cdot C/N$ and $\underline{\eps}(0) \geq 0$. \qedstar \\
\end{theorem}

\begin{proof}
See Appendix \ref{sec:proof_thm_explicit}.
\end{proof}

In \eqref{eq:epsU<=} and \eqref{eq:epsU>=}, the dependence in $\beta$ is inversely logarithmic, which shows that ``confidence is cheap''. For any fixed $k$, we see that $\overline{\eps}(k)$ and $\underline{\eps}(k)$ merge onto the same value $k/N$ as fast as $O(1/N)$, while for $k$ that grows at the same rate as $N$, say $k = \mu N$, convergence towards $k/N$ takes place at a rate $O(\ln (N)/\sqrt{N})$. Hence, we see that we can construct a strip around $k/N$ whose size goes to zero as $O(\ln (N)/\sqrt{N})$ and the bi-variate distribution of risk and complexity all lies in the strip but a slim tail that expands beyond the strip whose probability is no more than $\beta$.

Going back to the comparison between \eqref{eq:eps_GHS-T}, \eqref{eq:V-bound-GHST} and Theorem \ref{th:concentration-VS-relaxation}, this time in relation to asymptotic results, suppose first that $r^\ast = 0$ with probability one (no constraints violation - realizable case), in which case it holds that $s^\ast  = c^\ast$ with probability one. This allows for a simple comparison between $\tilde{\eps}(c,0)$ and the upper bound for $\overline{\eps}(k)$ in equation \eqref{eq:epsU<=}, from which the following conclusions can be drawn:
\begin{itemize}
	\item[(i)] for any given $\beta$ and for fixed value of $c$ and $k$, we have that both $\tilde{\eps}(c,0)$ and $\overline{\eps}(k)$ converge to $0$ as $N \to \infty$. Hence, both results capture the fact that the risk of $x^\ast$ goes to $0$ asymptotically when the complexity $s^\ast$ keeps bounded as the number $N$ of observations increases. The rate of convergence is slightly different and $\tilde{\eps}(c,0)$ converges to $0$ as $O(\ln(N)/N)$ while $\overline{\eps}(k)$ converges to zero as $O(1/N)$. Provably, $O(1/N)$ is the fastest possible rate of convergence, \cite{HannekeKontorovich_2019}. It is worth mentioning that a bound converging to $0$ as $O(1/N)$ was also obtained in \cite{pmlr-v125-bousquet20a}. This bound, however, differently from $\tilde{\eps}(r,0)$ and $\overline{\eps}(k)$ is only valid under the assumption that $s^\ast$ is uniformly upper bounded for any $N$;
	\item[(ii)] when $c$ and $k$ grow at the same rate as $N$, say $c = k = \mu N$, the behavior of $\overline{\eps}(k)$ and
	$\tilde{\eps}(c,0)$ are quite different. $\overline{\eps}(k)$ (and also $\underline{\eps}(k)$) converge to $k/N = \mu$ as $N \to \infty$ at a rate $O(\ln (N)/\sqrt{N})$. Instead, $\tilde{\eps}(c,0)$ does not converge to $c/N = \mu$ as $N \to \infty$, since, as shown in Appendix \ref{appendix:lower_bound_eps_tilde}, asymptotically it holds that $\tilde{\eps}(c,0) \geq 1-(1-c/N)(c/N)^{\frac{c/N}{1-c/N}}$ (which is bigger than $c/N = \mu$). This substantially different behavior can be also appreciated in Figure \ref{fig:eps_tilde_vs_eps}, where $\tilde{\eps}(c,0)$ along with $1-(1-c/N)(c/N)^{\frac{c/N}{1-c/N}}$ and $\overline{\eps}(k)$ and $\underline{\eps}(k)$ along with $k/N$ are plotted as functions of $c = 0,1,\ldots,N$ and of $k=0,1\ldots,N$ for increasing values of $N$. Here, we see that the actual profile of $\tilde{\eps}(c,0)$ departs significantly above $1-(1-c/N)(c/N)^{\frac{c/N}{1-c/N}}$ for large values of $c$ (in fact the only reason for pointing out the lower limit $1-(1-c/N)(c/N)^{\frac{c/N}{1-c/N}}$ was to show that convergence to $c/N$ was missing).
	\begin{figure}
		\centering
		\includegraphics[width=0.95\columnwidth]{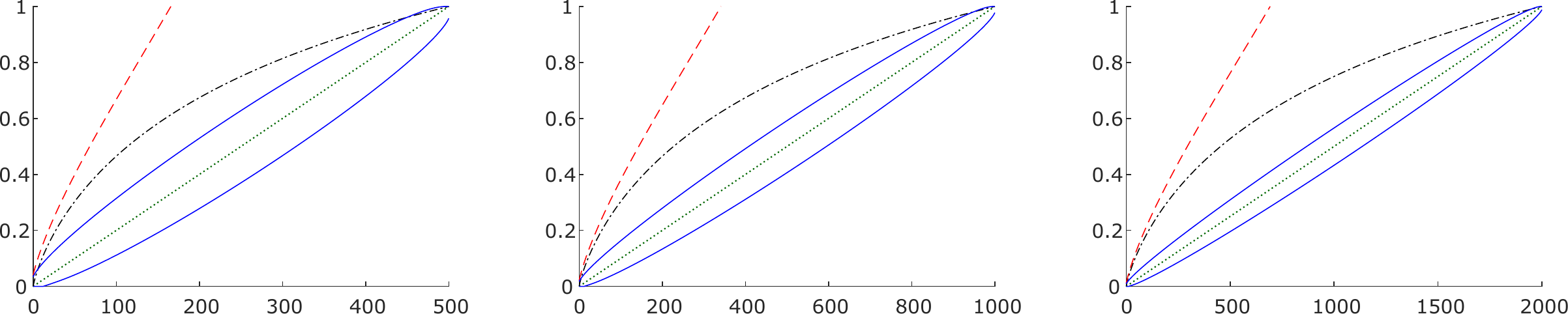}
		\caption{$\tilde{\eps}(c,0)$ (red dashed line), $1-(1-c/N)(c/N)^{\frac{c/N}{1-c/N}}$ (black dash-dotted line), $\overline{\eps}(k)$ and $\underline{\eps}(k)$ (blue solid lines), and $k/N$ (green dotted line) as functions of $c=0,1,\ldots,N$ and of $k=0,1\ldots,N$ for $N=500$, $N=1000$, and $N=2000$.}
		\label{fig:eps_tilde_vs_eps}
	\end{figure}
	The result that $1-\beta$ of the probabilistic mass of the bi-variate distribution of the risk and the complexity lies in the ``lenticular'' strip shown in Figure \ref{fig:eps_tilde_vs_eps} whose width shrinks to zero at a rate $O(\ln (N)/\sqrt{N})$	is unprecedented and shows for the first time that in all situations (including those where the complexity is not a-priori upper bounded, a condition that applies e.g. to Support Vector methods with kernels) the ratio $s^\ast/N$ is a consistent estimator of the risk $V(x^\ast)$.
\end{itemize}
For the case $r^\ast \neq 0$, similar to the discussion above one can conclude that $\tilde{\eps}(c,r)$ does not converge to $(c+r)/N$ in situations where $c$ grows unbounded. In contrast, in the theory of the present paper the fact that $r^\ast \neq 0$ is not explicitly considered as a separate case and the complexity $s^\ast$ accommodates all situations. Within this theory, the fact that for $k=\mu N$ both $\overline{\eps}(k)$ and $\underline{\eps}(k)$ converge to $k/N = \mu$ for all $\mu$ at a rate $O(\ln (N)/\sqrt{N})$ shows that $s^\ast$ is an observable by which $V(x^\ast)$ can be consistently estimated in all situations and it reveals that the distinction between $c^\ast$ and $r^\ast$ is fictitious for the purpose of evaluating the risk.

\section{Application to Support Vector Methods}
\label{sec:svm}

In this section, the general theory for optimization with constraints relaxation is applied to various well known support vector methods. The results stemming from this analysis are unprecedented and show that complexity carries fundamental information to tightly judge the ability of these machines to generalize.

We consider in turn: SVR (Support Vector Regression), SVDD (Support Vector Data Description) and SVM (Support Vector Machine). To SVR and SVDD the theoretical apparatus developed in the previous section can be directly applied, while SVM requires some additional effort to rigorously accommodate some degenerate situations; the analysis for SVM also shows the versatility of the theory.

\subsection{Support Vector Regression - SVR}
\label{subsection-SVR}
Let $\{\delta_i\}_{i=1}^N = \{(\mathbf{u}_i,y_i)\}_{i=1}^N$ be a data set, where the $\mathbf{u}_i$'s are elements of a Hilbert space $\mathcal{U}$ and the $y_i$'s are the corresponding output values in $\Real{}$. Each data point is drawn independently of the others from a common probability distribution $\prob$.

\begin{rmk}
	\label{remark-lifting}
	Depending on the application, values $\mathbf{u}_i$ can be thought of as raw measurements of physical quantities or rather as measurements lifted into a feature space by means of a feature map $\varphi(\cdot)$, so that $\mathbf{u}_i = \varphi(\mathbf{m}_i)$, where $\mathbf{m}_i$ is a vector of measured quantities. Interestingly, when SVR is applied, the actual computation of the solution involves only the evaluation of inner products in the feature space, that is, $\langle \varphi(\mathbf{m}_k),\varphi(\mathbf{m}_j)\rangle$, which can be done without explicitly evaluating $\varphi(\mathbf{m}_i)$. Indeed, one can define a ``kernel'' $k(\mathbf{m}_k,\mathbf{m}_j) := \langle \varphi(\mathbf{m}_k),\varphi(\mathbf{m}_j )\rangle$ and working with function $k(\cdot,\cdot)$ enables one to implicitly operate in the (high-dimensional) feature space without ever computing explicitly the coordinates of the measurements in the lifted feature space. This is the so-called ``kernel trick''. Pushing all this even further, it can be observed that for the operation of the method one does not even need to provide an explicit description of the inner product $\langle \cdot,\cdot \rangle$ and of the feature map $\varphi(\cdot)$ from which $k(\cdot,\cdot)$ is defined by composition: in fact one can start off by assigning $k(\cdot,\cdot)$ directly and theoretical results in RKHS -- Reproducing Kernel Hilbert Spaces -- assure that this always corresponds to allocate a suitable couple $\langle \cdot,\cdot \rangle$ and $\varphi(\cdot)$ so that $k(\cdot,\cdot) = \langle \varphi(\cdot),\varphi(\cdot) \rangle$, provided that the kernel is positive definite (i.e., $\sum_{i=1}^n \sum_{j=1}^n k(\mathbf{m}_i,\mathbf{m}_j) c_i c_j \geq 0$, for all choices of $n$ and all finite sequences of points $(\mathbf{m}_1, \ldots, \mathbf{m}_n)$ and real values $(c_1, \ldots, c_n)$). When adopting this standpoint, the interpretation of $k(\cdot,\cdot)$ is that it is a user-specified similarity function over pairs of data points in raw representation. \qedstar
\end{rmk}

In the following, we refer to SVR with adjustable size as described in \cite{ScBaSmWi:98}. For given parameters $\tau, \rho > 0$, consider the following optimization program, which, for easy reference, we repeat from the introduction (we here also specify more precisely the domain of optimization)\footnote{The above SVR formulation is suitable when the data are homoskedastic. Other convex formulations exist to model heteroskedastic processes, see \cite{crespo2016interval}.}
\begin{align}
\label{svr-1}
\min_{w \in \mathcal{U}, \gamma \geq 0, b \in \mathbb{R} \atop \xi_i \geq 0, i=1,\ldots,N} & \quad  (\gamma + \tau \| w \|^2) + \rho \sum_{i=1}^N \xi_i \\
\textrm{\rm subject to:} & \quad |y_i - \langle w,\mathbf{u}_i \rangle - b | - \gamma \leq \xi_i, \ \ i = 1, \ldots ,N. \nonumber
\end{align}
In this context the risk is interpreted as the probability of an erroneous prediction (i.e., a new observation $(\bar{u},\bar{y})$ is not in the tube \eqref{opt-tube} constructed by SVR). Also, notice that the size of the prediction tube is known from the solution of the optimization program, while the theory here developed provides a fundamental grasp on the other relevant quantity, the probability of an erroneous prediction. These two pieces of information form the beacon to select a suitable value of the tuning parameter $\rho$. See also Section \ref{sec:numerical_example} for a numerical example.
\begin{rmk}
	It is perhaps worth re-writing program \eqref{svr-1} in terms of the original measurements when one resorts to a kernel lifting. As we show below, $w^\ast$ is always given by a linear combination of the points $\mathbf{u}_i$, so that in \eqref{svr-1} one can only consider solutions of the type $w = \sum_j \alpha_j \mathbf{u}_j$. Thus, one obtains $\| w \|^2 = \langle \sum_j \alpha_j \mathbf{u}_j, \sum_k \alpha_k \mathbf{u}_k \rangle = \sum_{j,k} \alpha_j \alpha_k \langle  \mathbf{u}_j, \mathbf{u}_k \rangle$, while the constraints can be re-written as $|y_i - \sum_j \alpha_j \langle \mathbf{u}_j,\mathbf{u}_i \rangle - b | - \gamma \leq \xi_i$, $i = 1, \ldots,N$. Thus, in kernel notation, program \eqref{svr-1} becomes	
	\begin{align} \label{svr-2}
		\min_{\alpha_j, j=1,\ldots,N, \gamma \geq 0, b \in \mathbb{R} \atop \xi_i \geq 0, i=1,\ldots,N} & \quad  \gamma + \tau \sum_{j,k} \alpha_j \alpha_k k(\mathbf{m}_j, \mathbf{m}_k) + \rho \sum_{i=1}^N \xi_i \\
		\textrm{\rm subject to:} & \quad |y_i - \sum_j \alpha_j k(\mathbf{m}_j,\mathbf{m}_i) - b | - \gamma \leq \xi_i, \ \ i = 1, \ldots ,N, \nonumber
	\end{align}
	which is a quadratic program. Often, \eqref{svr-2} is solved by resorting to its dual formulation. Letting $\alpha_j^\ast$, $j=1,\ldots,N$, $\gamma^\ast$, $b^\ast$, and $\xi_i^\ast$, $i=1,\ldots,N$ be the optimal solution to \eqref{svr-2}, the tube in \eqref{opt-tube} becomes
	$$
	|y - \sum_j \alpha_j^\ast k(\mathbf{m}_j,\mathbf{m}) - b^\ast | \leq \gamma^\ast.
	$$
	\hfill \qedstar
\end{rmk}
Throughout, we make the following assumption.
\\
\begin{assumption}
	Over the support of $\mathbf{u}$, the conditional distribution of $y$ given $\mathbf{u}$ admits density. \qedstar \\
\end{assumption}

In order to apply the theory from Section \ref{sec:theory} we need to show that the solution to \eqref{svr-1} exists and is unique (Assumption \ref{asmpt:existence-uniqueness-generalized}) and that a non-accumulation assumption applies (Assumption \ref{asmpt: f concentration}). The validity of these facts is shown in the following. \\
\\
\emph{Existence}: While $w$ belongs to a possibly infinite dimensional Hilbert space $\mathcal{U}$, the minimization problem in \eqref{svr-1} (with $m$ in place of $N$ as required in Assumption \ref{asmpt:existence-uniqueness-generalized}) can be seen as finite dimensional because allowing for components of $w$ outside the finite dimensional span of points $\mathbf{u}_i$, $i = 1, \ldots, m$, does not help satisfy the constraints (note that in the constraints $w$ shows up only under the sign of inner product $\langle w,\mathbf{u}_i \rangle$), while it increases the cost function (write $w = w_\mathbf{u} + w_\mathbf{u}^\perp$, with $w_\mathbf{u} \in$ span of $\mathbf{u}_i$, $i = 1, \ldots, m$, and $w_\mathbf{u}^\perp$ orthogonal to the same span, and then apply Pitagora's theorem: $\|w\|^2 = \|w_\mathbf{u}\|^2 + \|w_\mathbf{u}^\perp\|^2$). Hence, \eqref{svr-1} is a finite-dimensional problem with closed constraints and quadratic non-negative cost over the optimization domain. As such, it  certainly admits solution. \\
\\
\emph{Uniqueness}: At optimum, $w^\ast$ is certainly unique because, assuming by contradiction that there are two optimal solutions $(w_1^\ast,\gamma_1^\ast,b_1^\ast,\xi_{i,1}^\ast)$ and $(w_2^\ast,\gamma_2^\ast,b_2^\ast,\xi_{i,2}^\ast)$ with $w_1^\ast \neq w_2^\ast$, then an easy computation shows that the point half way between these two solutions would be feasible and superoptimal (the reader may also want to refer to Theorem 3 in \cite{BurgesCrisp:99} where the same issue is discussed in relation to an algorithmically  slightly different, but conceptually identical, problem). Instead, $\gamma^\ast$, $b^\ast$ and $\xi^\ast_i$ might be non-unique. To identify a unique solution we select the smallest $\gamma^\ast$ and the $b^\ast$ with smallest absolute value. Note that this certainly breaks the tie because the smallest $\gamma^\ast$ is obviously unique while, if one had two values for $b^\ast$ smallest in absolute value, say $b^\ast = \pm \bar{b}$, corresponding to the solutions $(w^\ast,\gamma^\ast,\bar{b},\xi_{i,1}^\ast)$ and $(w^\ast,\gamma^\ast,-\bar{b},\xi_{i,2}^\ast)$ (recall that $w^\ast$ and $\gamma^\ast$ must be the same at optimum), then optimality of these two solutions would imply that $\sum_{i=1}^N \xi_{i,1}^\ast = \sum_{i=1}^N \xi_{i,2}^\ast$ and therefore the solution half way between $(w^\ast,\gamma^\ast,\bar{b},\xi_{i,1})$ and $(w^\ast,\gamma^\ast,-\bar{b},\xi_{i,2})$, i.e., $(w^\ast,\gamma^\ast,0,0.5 \cdot \xi_{i,1} + 0.5 \cdot \xi_{i,2})$, would be feasible thanks to convexity, it would achieve the same cost as the other two solutions, but it would be preferred because it carries a smaller value for $|b^\ast|$ than in the two alleged solutions. Once $w^\ast$, $\gamma^\ast$ and $b^\ast$ are uniquely determined, also the $\xi^\ast_i$'s remain determined, see the footnote at the end of Assumption \ref{asmpt:existence-uniqueness-generalized}. \\
\\
\emph{Non-accumulation}: Non-accumulation requires that, $\forall w, \gamma , b$, one has:
$$
\prob \{ | y - \langle w,\mathbf{u} \rangle - b | - \gamma = 0 \} = 0.
$$
Since the conditional distribution of $y$ given $\mathbf{u}$ admits density, one has $\prob \{ | y - \langle w,\mathbf{u} \rangle - b | - \gamma = 0 \} = \prob \{ \prob\{ | y - \langle w,\mathbf{u} \rangle - b | - \gamma = 0 | \mathbf{u} \} \} = \prob \{ \prob\{ y = \langle w,\mathbf{u} \rangle + b \pm \gamma | \mathbf{u} \} \} = 0$. \\

Since all conditions are satisfied, we can apply Theorem \ref{th:concentration-VS-relaxation} to SVR, which gives the following result. \\

\begin{theorem}[Reliability of SVR]
	\label{theorem-SVR}
	With $\underline{\eps}(\cdot)$ and $\overline{\eps}(\cdot)$ as defined in Theorem \ref{th:concentration-VS-relaxation}, we have
	$$
	\prob^N \{ \underline{\eps}(s^\ast) \leq \prob\{ (\mathbf{u},y): |y - \langle w^\ast,\mathbf{u} \rangle - b^\ast | > \gamma^\ast \}
	\leq \overline{\eps}(s^\ast) \} \geq 1 - \beta,
	$$
	where $s^\ast$ is the number of $(\mathbf{u}_i,y_i)$'s for which $|y_i - \langle w^\ast,\mathbf{u_i} \rangle - b^\ast | \geq \gamma^\ast$. \qedstar
\end{theorem}

\subsection{Support Vector Data Description - SVDD}

Support Vector Data Description is a data-driven technique used to identify a portion of space that covers most of the probabilistic mass from which data have been generated, while including little superfluous space. SVDD creates a spherically shaped form and, analogous to SVR, it can be made more flexible by lifting the data into a feature space so as to obtain more complex geometries in the original measurement space. See e.g. \cite{TaxDuin2004} for a more comprehensive description. See also \cite{CreColKenGie2019} for an approach that allows one to apply SVDD to more complex geometries when  working directly in the measurement space.

Let $\{\delta_i\}_{i=1}^N = \{\mathbf{p}_i\}_{i=1}^N$ be an independent data set in a Hilbert space $\mathcal{P}$ drawn from a common probability distribution $\prob$. These points can be raw data or, in complete analogy with the discussion in Remark \ref{remark-lifting}, data lifted into a feature space by means of a map $\varphi(\cdot)$. SVDD constructs a sphere in $\mathcal{P}$ by solving the following optimization program:
\begin{align}
\label{svdd}
\min_{c \in \mathcal{P},\gamma \geq 0 \atop \xi_i \geq 0, i=1,\ldots,N} & \quad  \gamma + \rho \sum_{i=1}^N \xi_i \\
\textrm{\rm subject to:} & \quad \| \mathbf{p}_i - c \|^2 - \gamma \leq \xi_i, \ \ i = 1, \ldots ,N. \nonumber
\end{align}
\begin{rmk}
When the original data are lifted into a feature space ($\mathbf{p} = \varphi(\mathbf{m})$) defined through a kernel, considering that the solution takes the form $c = \sum_j \alpha_j \mathbf{p_j}$ (see below), program \eqref{svdd} can be re-written as 
\begin{align}
	\label{svdd-1}
	\min_{\alpha_j, j=1,\ldots,N,\gamma \geq 0 \atop \xi_i \geq 0, i=1,\ldots,N} & \quad  \gamma + \rho \sum_{i=1}^N \xi_i \\
	\textrm{\rm subject to:} & \quad k(\mathbf{m}_i,\mathbf{m}_i) + \sum_{j,k} \alpha_j \alpha_k k(\mathbf{m}_j,\mathbf{m}_k)  - 2 \sum_{j} \alpha_j k(\mathbf{m}_i,\mathbf{m}_j)  - \gamma \leq \xi_i, \ \ i = 1, \ldots ,N. \nonumber
\end{align}
Moreover, the region in the measurement space obtained by the optimization procedure is given by 
$$
\Big\{ \mathbf{m}: \quad k(\mathbf{m},\mathbf{m}) + \sum_{j,k} \alpha_j^\ast \alpha_k^\ast k(\mathbf{m}_j,\mathbf{m}_k)  - 2 \sum_{j} \alpha_j^\ast k(\mathbf{m},\mathbf{m}_j)  \leq \gamma^\ast \Big\}.
$$ \qedstar
\end{rmk}

We next address existence, uniqueness and non-accumulation for this problem. \\
\\
\emph{Existence}: Similarly to SVR, the optimal $c^\ast$ must belong to the finite dimensional space generated by $\mathbf{p}_i$, $i = 1, \ldots, m$, and a solution to \eqref{svdd} certainly exists. \\
\\
\emph{Uniqueness}: At optimum, the center of the sphere $c^\ast$ is unique while $\gamma^\ast$ and the $\xi_i^\ast$'s may not be unique, refer to Theorems 2 and 3 in \cite{WangChungWang2011}; moreover non-uniqueness may only occur when $\rho =  1/M$ for some integer $M$, refer again to Theorem 3 in \cite{WangChungWang2011}. To break the tie if it occurs, select the smallest $\gamma^\ast$; note that in this way also the $\xi_i^\ast$'s remain uniquely determined as explained in the footnote at the end of Assumption \ref{asmpt:existence-uniqueness-generalized}. \\
\\
\emph{Non-accumulation}: For SVDD, non-accumulation requires the following condition to hold.
\begin{assumption}
	For any $c$ and $\gamma$ it holds that
	\begin{equation}
	\label{non-degeneracy-SVDD}
	\prob \{  \| \mathbf{p} - c \|^2 = \gamma \} = 0.
	\end{equation} \qedstar
\end{assumption}
This condition simply requires that probabilistic mass does not accumulate over hyper-spheres. \\

We now have the following theorem. \\

\begin{theorem}[Reliability of SVDD]
	With $\underline{\eps}(\cdot)$ and $\overline{\eps}(\cdot)$ as defined in Theorem \ref{th:concentration-VS-relaxation}, we have
	$$
	\prob^N \{ \underline{\eps}(s^\ast) \leq \prob\{ \mathbf{p}: \| \mathbf{p} - c^\ast \|^2 > \gamma^\ast \} \leq \overline{\eps}(s^\ast) \}
	\geq 1 - \beta,
	$$
	where $s^\ast$ is the number of $\mathbf{p}_i$'s for which $\| \mathbf{p}_i - c^\ast \|^2 \geq \gamma^\ast$.  \qedstar
\end{theorem}

\subsection{Support Vector Machines - SVM}

SVM is a well-known technique that constructs binary classifiers from a data set. Given a new out-of-sample case, the classifier predicts its label to be $-1$ or $1$. $-1$ and $1$ represent two different classes, whose meaning depends on the application at hand and can e.g. be \emph{sick} or \emph{healthy}, \emph{right} or \emph{wrong}, \emph{male} or \emph{female}. Among the vast literature on SVM, refer e.g. to \cite{CortesVapnik1995,SchSmola_BOOK}.

Let $\{\delta_i\}_{i=1}^N = \{(\mathbf{u}_i,y_i)\}_{i=1}^N$ be a data set of independent observations from a common probability distribution $\prob$, where the $\mathbf{u}_i$'s are elements of a Hilbert space $\mathcal{U}$ and the $y_i$'s are the corresponding labels, $-1$ or $1$. Similarly to SVR, the $\mathbf{u}_i$'s can be thought of as raw measurements or measurements lifted into a feature space, refer to Remark \ref{remark-lifting}.

The classifier is obtained by solving the program:
\begin{align}
\label{svm}
\min_{w \in \mathcal{U}, b \in \mathbb{R} \atop \xi_i \geq 0, i=1,\ldots,N} & \quad  \| w \|^2 + \rho \sum_{i=1}^N \xi_i \\
\textrm{\rm subject to:} & \quad 1 - y_i (\langle w,\mathbf{u}_i \rangle - b) \leq \xi_i, \ \ i = 1, \ldots ,N, \nonumber
\end{align}
which gives the classifier $\hat y = \sign(\langle w^\ast,\mathbf{u} \rangle - b^\ast)$ (``$\ast$'' denotes the solution to \eqref{svm}).

\begin{rmk}
In case of lifting into a feature space, considering that the solution takes the form $w = \sum_j \alpha_j \mathbf{u}_j$ (see below), program \eqref{svm} can be re-written as 
\begin{align*} 
\min_{\alpha_j, j=1,\ldots,N, b \in \mathbb{R} \atop \xi_i \geq 0, i=1,\ldots,N} & \quad  \sum_{j,k} \alpha_j \alpha_k k(\mathbf{m}_j, \mathbf{m}_k) + \rho \sum_{i=1}^N \xi_i \\
\textrm{\rm subject to:} & \quad 1 - y_i \left( \sum_j \alpha_j k(\mathbf{m}_j,\mathbf{m}_i) - b\right) \leq \xi_i, \ \ i = 1, \ldots ,N, \nonumber
\end{align*}
with the classifier given by $\hat y = \sign( \sum_j \alpha_j^\ast k(\mathbf{m}_j,\mathbf{m}) - b^\ast)$. \qedstar
\end{rmk}

Existence and uniqueness of the solution $(w^\ast,b^\ast,\xi_i^\ast)$ present no difficulties. In contrast, non-accumulation raises some subtle issues (which refer to the situation where $w^\ast = 0$) that make a rigorous application of the results from Section \ref{sec:theory} non-trivial. \\
\\
\emph{Existence}: As in previous support vector methods, $w^\ast$ must belong to a finite dimensional subspace spanned by $\{ \mathbf{u}_i, i=1,\ldots,N \}$ and an optimal solution certainly exists. \\
\\
\emph{Uniqueness}: $w^\ast$ is unique while $b^\ast$ may not be, see Theorem 2 in \cite{BurgesCrisp:99}. Break the tie by minimizing $|b+1|$.\footnote{The reason for choosing $|b+1|$ and not $|b|$ is that this prevents the solution $w^\ast = 0$ and $b^\ast = 0$ from happening (which would result in a not-well defined classifier, see below).} Similarly to SVR, this returns unique $w^\ast$ and $b^\ast$ and the $\xi^\ast_i$'s also remain uniquely determined, see Remark \ref{remark-lifting}.\\
\\
\emph{Non-accumulation}: It requires satisfaction of the condition
$$
\prob \{1 - y (\langle w,\mathbf{u} \rangle - b) = 0 \} = 0 \quad \forall w,b.
$$
A problem with this condition rises for $w = 0$ and $b = \pm 1$, in which case the condition becomes
$$
\prob \{1 \pm y = 0\} = 0,
$$
which is generally not satisfied. This is sign of an intrinsic difficulty: if one sees all labels of one type $-1$ or $1$ (which happens with nonzero probability), then program \eqref{svm} returns $w^\ast =0$ and $-b^\ast = 1$ (in case of all labels equal to $1$) or $-b^\ast = -1$ (in case of all labels equal to $-1)$. Then, one ends up in a \emph{degenerate} situation where the solution is identified by various subsets of the data set (think of when all labels are $1$: any non-empty subset of data points returns the same solution), which is exactly what the non-accumulation Assumption \ref{asmpt: f concentration} rules out. Moreover, seeing all labels of one type is not the only case in which $w^\ast =0$ and $b^\ast = \pm 1$ and it is easy to figure out other configurations of data points for this to happen. In all these cases, degeneracy occurs. Hence, the fact that the non-accumulation Assumption \ref{asmpt: f concentration} is not satisfied is not accidental and has deep motivations. Nevertheless, we can get around this difficulty and get the theory to work for a \emph{heated} version of the problem. By a \emph{cooling} procedure, one then finds rigorous results for SVM. Along this route, we also introduce a breakdown of the initial optimization problem into three distinct problems where a problem that has a specific  simple structure is considered when one knows that $w^\ast = 0$ for the initial problem \eqref{svm}; this is instrumental to finding tight evaluations of the risk for this case as well. One side effect of this process is that in the final result the confidence parameter is elevated from the value $\beta$ to the value $3\beta$, which has however very little impact in practice. The technically articulated theory is presented in Appendix \ref{sec:proof_thm_vio_SVM}, while we give here the final result. The result requires that $\mathbf{u}$ are generically distributed and do not concentrate on linear manifolds, as the following assumption states.

\begin{assumption} \label{non-degeneracy-SVM}
	Assume that
	$$\prob \{ (\mathbf{u},y): \langle a,\mathbf{u} \rangle - h = 0\} = 0 \quad \forall a \neq 0, h.
	$$
	\qedstar \\
\end{assumption}

\begin{theorem}[Violation of SVM]
	\label{theorem-violation-SVM}
	With $\underline{\eps}(\cdot)$ and $\overline{\eps}(\cdot)$ as defined in Theorem \ref{th:concentration-VS-relaxation}, we have
	$$
	\prob^N \{ \underline{\eps}(s^\ast) \leq \prob\{ (\mathbf{u},y): 1 - y (\langle w^\ast,\mathbf{u} \rangle - b^\ast) > 0\} \leq \overline{\eps}(s^\ast) \} \geq 1 - 3\beta,
	$$
	where $s^\ast$ is so defined: when $w^\ast \neq 0$, $s^\ast$ is the number of $(\mathbf{u}_i,y_i)$'s for which $1 - y_i (\langle w^\ast,\mathbf{u}_i \rangle - b^\ast) \geq 0$ and, when $w^\ast = 0$, $s^\ast$ is the number of data points whose label belongs to the class with fewer elements (if, e.g., there are $960$ data points with label $1$ and $40$ with label $-1$, then $s^\ast = 40$; if there is a fifty-fifty split, then $s^\ast$ is equal to half of the data points). \qedstar \\
\end{theorem}

\begin{proof}
See Appendix \ref{sec:proof_thm_vio_SVM}.
\end{proof}

One further point that needs be clearly highlighted is that in SVM constraints violation does not correspond to misclassification. This marks a difference with SVR and SVDD where indeed constraints violation meant misprediction and was the final quantity that we wanted to keep under control. To understand this point, refer to the classifier generated by SVM:
\begin{eqnarray}
& & \mbox{classify as } 1 \mbox { points } \mathbf{u} \mbox{ such that } \langle w^\ast,\mathbf{u} \rangle - b^\ast > 0;  \nonumber \\
& & \mbox{classify as } -1 \mbox { points } \mathbf{u} \mbox{ such that } \langle w^\ast,\mathbf{u} \rangle - b^\ast < 0. \nonumber
\end{eqnarray}
Hence, we make an error if $(\mathbf{u},y)$ is such that
\begin{equation}
\label{error-SVM}
y(\langle w^\ast,\mathbf{u} \rangle - b^\ast) < 0,
\end{equation}
corresponding to having disagreement between the classifier and the actual sign of $y$. This condition is more restrictive than constraints violation, and in fact \eqref{error-SVM} implies (and is not implied by)
$$
1 - y (\langle w^\ast,\mathbf{u} \rangle - b^\ast) > 0.
$$
Hence, misclassification occurs more rarely than constraints violation. As a consequence, Theorem \ref{theorem-violation-SVM} can be used to only upper bound the probability of misclassification, a result that is stated in the next theorem. \\

\begin{theorem}[Misclassification of SVM]
	\label{theorem-misclassification-SVR}
	Define $\overline{\eps}(\cdot)$ as in Theorem \ref{th:concentration-VS-relaxation}. We have
	\begin{equation}
	\prob^N \{ \prob\{ (\mathbf{u},y):
	y \mbox{ is misclassified} \}
	\leq \overline{\eps}(s^\ast) \}
	\geq 1 - 3\beta,
	\end{equation}
	where $s^\ast$ is so defined: when $w^\ast \neq 0$, $s^\ast$ is the number of $(\mathbf{u}_i,y_i)$'s for which $1 - y_i (\langle w^\ast,\mathbf{u}_i \rangle - b^\ast) \geq 0$ and, when $w^\ast = 0$, $s^\ast$ is the number of data points whose label belongs to the class with fewer elements. \qedstar \\
\end{theorem}

\begin{rmk}[Sensitivity and specificity] The probability of misclassification $\prob\{ (\mathbf{u},y): y \mbox{ is misclassified} \}$ is also known in the machine learning literature as ``accuracy''. Depending on the application at hand, it may be that the user is also interested in the so-called ``specificity'' and ``sensitivity'', which are the probability of misclassification within a given class, either $1$ or $-1$ (in formulas:	specificity  =  $\prob\{ (\mathbf{u},y): y \text{ is misclassified} \; | \; y = 1\}$; sensitivity = $\prob\{ (\mathbf{u},y): y \mbox{ is misclassified} \; | \; y = -1\}$). The theory that has been presented here can be easily modified to also provide a characterization of specificity and sensitivity, see \cite{CarRamCam2018} for a similar argument applied to a different context. We sketch in this remark how this is achieved. Consider e.g. specificity. Let $N^\ast_1$ be the number of observations, among the $N$ available, for which $y_i = 1$. Take conditioning on the value of $N^\ast_1$ and on the value of the remaining $N-N^\ast_1$ observations for which $y_i = -1$. The $N^\ast_1$ observations with $y_i = 1$ are instead let vary and seen as random. With the definition that $\tilde{\prob}_1$ is the probability over $\mathcal{U}\times\{-1,1\}$ obtained after conditioning on $y = 1$ (that is, $\tilde{\prob}_1(E) = \prob(E \; | \; y=1)$ for every measurable event $E \subseteq \mathcal{U}\times\{-1,1\}$), the argument in the proof of Theorems \ref{theorem-violation-SVM} and \ref{theorem-misclassification-SVR} can be repeated \emph{mutatis mutandis} to obtain
	\begin{equation} \label{eq:prob_sensitivity_aux}
	\tilde{\prob}_1^{N^\ast_1} \{ \tilde{\prob}_1 \{ (\mathbf{u},y): y \mbox{ is misclassified} \} \leq \overline{\eps}_{N^\ast_{1}}(s^\ast_1) \} \geq 1 - 3\beta,
	\end{equation}
	where $\overline{\eps}_{N^\ast_1}$ is the same function as $\overline{\eps}$ (calculated for $N^\ast_1$ observations) and $s^\ast_1$ is defined similarly to $s^\ast$ in Theorem \ref{theorem-violation-SVM} but limited to observations for which $y_i = 1$ (in the proof, observations $y_i = -1$ have to be thought of as fixed and they do not concur in the evaluation of the complexity). Next, integrating the result in \eqref{eq:prob_sensitivity_aux} over the value of $N^\ast_1$ and over the variability of the observations with $y_i = -1$ yields
	$$
	\prob^N \{ \prob\{ (\mathbf{u},y): y \mbox{ is misclassified} \; | \; y = 1 \} \leq \overline{\eps}_{N^\ast_{1}}(s^\ast_1) \} \geq 1 - 3\beta,
	$$
	which is the characterization of specificity. Sensitivity is dealt with analogously. \qedstar
\end{rmk}

\section{Numerical Example} \label{sec:numerical_example}

\begin{figure}
	\centering
	\includegraphics[width=0.6\columnwidth]{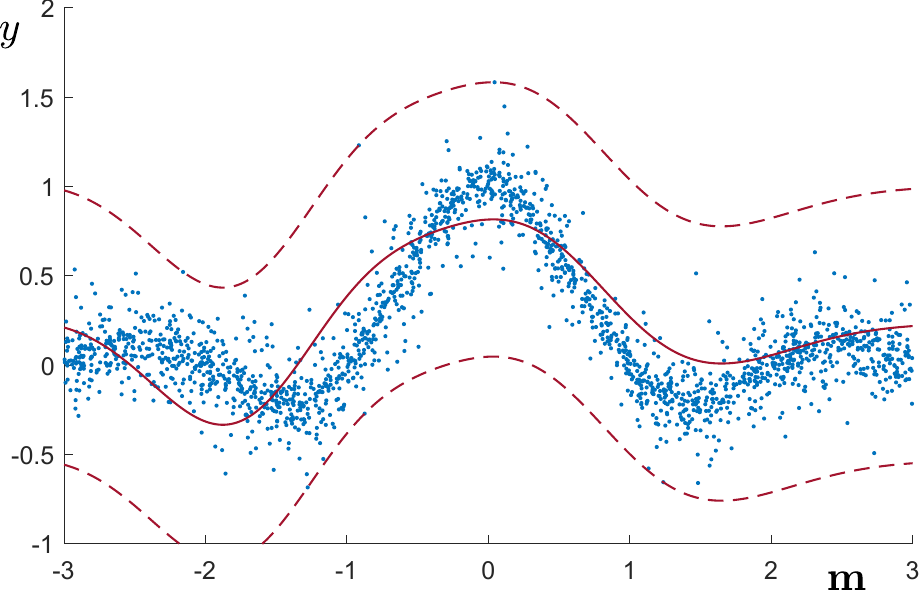}
	\caption{SVR model for $\rho = 1$.}
	\label{fig:svr_1}
\end{figure}

Inspired by the numerical example in \cite{ScBaSmWi:98}, we applied SVR to find a regression model for points generated by a noisy $\mathrm{sinc}$ function. Specifically, we considered a data set formed by $N=2000$ examples $(\mathbf{m}_i,y_i)$ with $\mathbf{m}_i$ extracted uniformly from $[-3,3]$ and $y_i = \sin(\pi \mathbf{m}_i)/(\pi \mathbf{m}_i) + e_i$, where $e_i$ had a Laplace distribution with mean $\mu=0$ and parameter $b=1$ (data points are in Figure \ref{fig:svr_1}). We used the Gaussian kernel $\exp(-|\mathbf{m}_k - \mathbf{m}_j|^2/\sigma^2)$, where $\sigma$ was regarded as an adjustable hyper-parameter.\footnote{Note that by means of parameter $\sigma$ one can tune the locality of the basis functions associated to the Gaussian kernel and selecting a small $\sigma$ corresponds to fine-grained basis functions with better descriptive capabilities (tantamount to what is achieved with polynomial kernels by increasing their order). Thus, small $\sigma$'s correspond to improved cost values, but also to increased solution complexities and, therefore, to higher risks.} With $\tau = 0.01$, program \eqref{svr-1} was repeatedly solved with $\sigma = 10^{k}$, $k=-2,-1,0,1$ and $\rho = (3/5)^\ell$, $\ell=0,1,\ldots,14$. Each time, we recorded: the solution; the cost; the value of the complexity $s^\ast$; and the interval for the risk $[\underline{\eps}(s^\ast),\overline{\eps}(s^\ast)]$ with $\beta=10^{-4}$, which was calculated as indicated in Theorem \ref{theorem-SVR}.

\begin{figure}[h!]
	\centering
	\includegraphics[width=0.7\columnwidth]{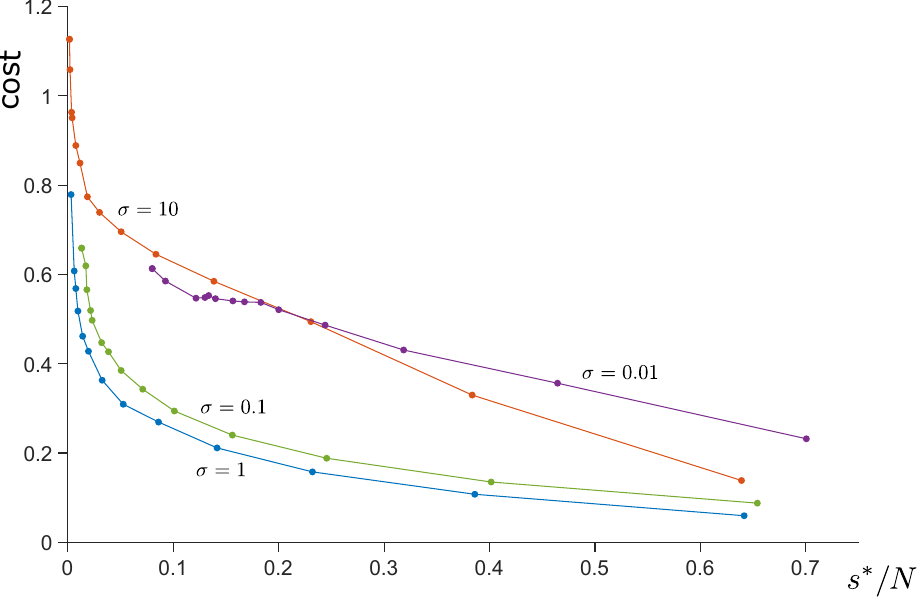}
	\caption{cost vs. value of $s^\ast/N$ for the various solutions.}
	\label{fig:costs_sig}
\end{figure}
Figure \ref{fig:costs_sig} shows the results: the $y$-axis gives the value of the cost while the $x$-axis contains the value of $s^\ast/N$ (which is an indicator of the risk). From this figure, it appears that $\sigma = 1$ dominates over other choices of $\sigma$.

Focusing on the solutions obtained for $\sigma=1$ we then constructed a cost-risk plot putting on display the cost and the corresponding interval for the risk ($\beta = 10^{-4}$) obtained for various values of $\rho$. The plot is that of Figure \ref{fig:cost-risk}. For $\rho = 1$ we obtained the smallest value for $s^\ast$, whence the range for the risk hit its minimum, at the expense of a large cost. As $\rho$ was decreased, $s^\ast$ showed a monotonic growth. Initially, the drop in the cost was rapid, paired with a moderate increase of the risk. Instead, for smaller values of $\rho$, even a small decrease of cost implied a significant rise of the risk. We opted for $\bar{\rho} = (3/5)^9$, yielding $s^\ast = 105$ (corresponding to $[\underline{\eps}(105),\overline{\eps}(105)] = [0.032,0.08]$) and $c(x^\ast) = 0.31$. Since $\tau$ was small, the cost $c(x^\ast)$ is almost identical to $\gamma^\ast$, the size of the tube.
\begin{figure}
	\centering
	\includegraphics[width=0.6\columnwidth]{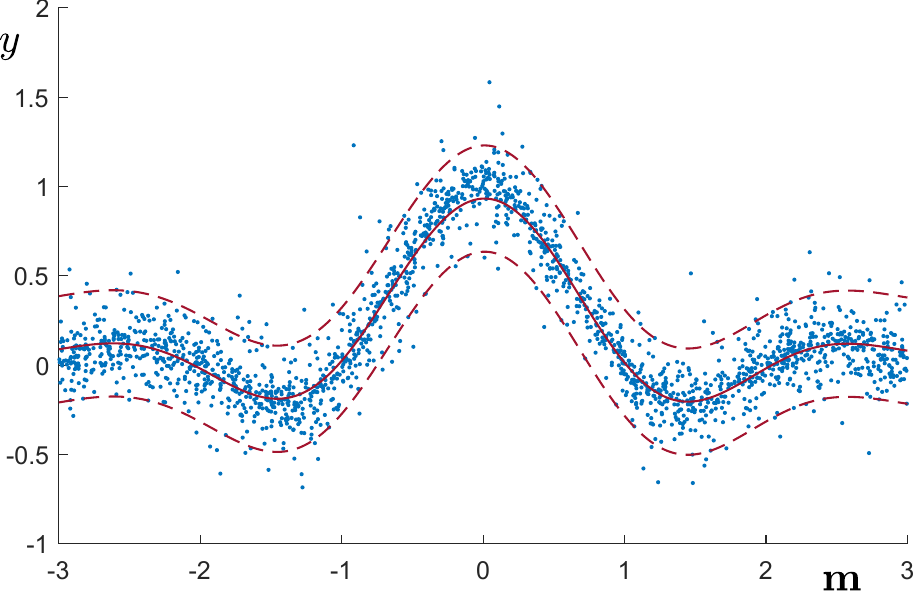}
	\caption{SVR model for $\rho = (3/5)^{9}$.}
	\label{fig:svr_2}
\end{figure}
\begin{figure}
	\centering
	\includegraphics[width=0.6\columnwidth]{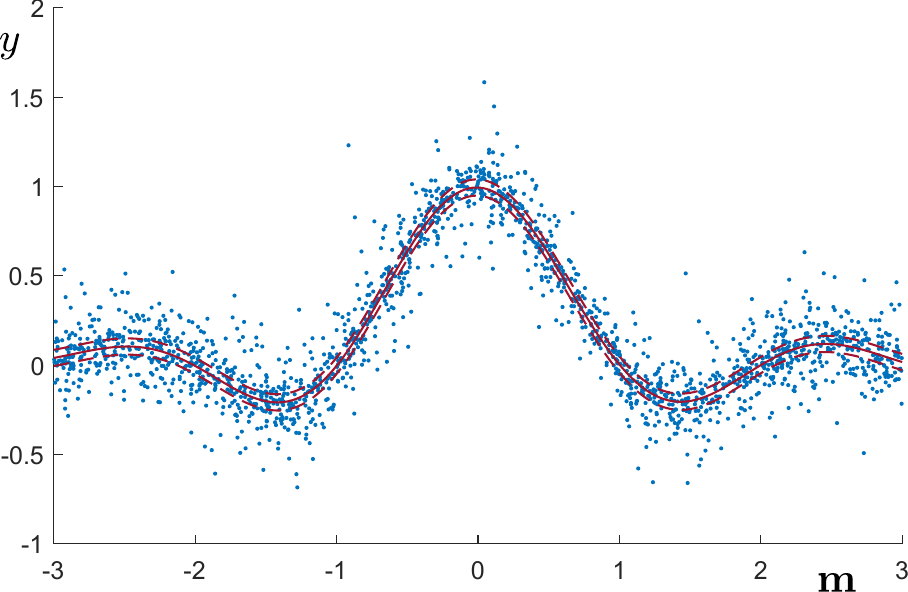}
	\caption{SVR model for $\rho = (3/5)^{14}$.}
	\label{fig:svr_4}
\end{figure}
Figures \ref{fig:svr_1}, \ref{fig:svr_2}, and \ref{fig:svr_4} depict the models obtained for $\rho = 1$, $\rho = (3/5)^9$ (our choice), and $\rho = (3/5)^{14}$. A visual inspection, possible in this case because we are considering a toy example with $\mathbf{m}$ scalar, confirms the analysis based on the ground of the cost-risk plot.

Finally, we thought it might be useful to look closer at the validity and tightness of Theorem \ref{th:concentration-VS-relaxation}. While keeping $\rho$ at the value $(3/5)^9$, we solved problem \eqref{svr-1} $200$ times, each time drawing a new sample of size $2000$. Each solution was then tested on $10000$ additional random values of $(\mathbf{m},y)$ to evaluate its risk (Monte-Carlo approach).
\begin{figure}
	\centering
	\includegraphics[width=0.6\columnwidth]{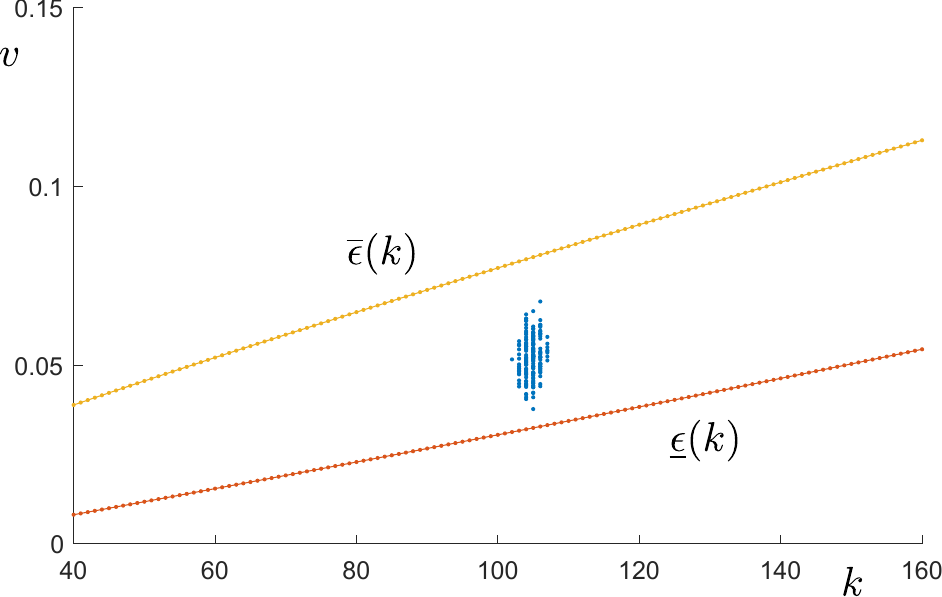}
	\caption{(complexity,risks) pairs (blue dots) vs. $\underline{\eps}(k)$ and $\overline{\eps}(k)$ (continuous dotted lines); $N = 2000$ and $\beta = 10^{-4}$.}
	\label{fig:epsLUvsVIO}
\end{figure}
Figure \ref{fig:epsLUvsVIO} plots the pairs (complexity,risk) obtained in the $200$ trials, along with the upper and lower limits $\underline{\eps}(k)$ and $\overline{\eps}(k)$ for $\beta = 10^{-4}$. Theorem \ref{th:concentration-VS-relaxation} predicts that the risk is, on average, in the interval $\underline{\eps}(k)$ and $\overline{\eps}(k)$ in $9999$ times out of $10000$. This was the case for all the $200$ points in our simulations. A visual inspection also reveals that the spread of the risks fills well the vertical range given by the theoretical bounds, a sign that the theoretical result provides tight evaluations in spite of its prerogative of being distribution-free.

\acks{The authors are indebtedd to Nicolò Cesa-Bianchi and Steve Hanneke for insightful comments and discussion on the content of this paper.}



\appendix

\section{Proof of Theorem \ref{th:explicit}} \label{sec:proof_thm_explicit}

Let $v := 1-t$. Equation \eqref{pol_eq-for-eps(k)-relax} for $k = 0,\ldots,N-1$ becomes
\begin{equation}
\label{eq:epsLU_eq}
\frac{\beta}{2N} \sum_{i=k}^{N-1} {i \choose k}(1-v)^{i-k} + \frac{\beta}{6N} \sum_{i=N+1}^{4N} {i \choose k}(1-v)^{i-k}
= {N \choose k}(1-v)^{N-k}.
\end{equation}
The fact that \eqref{pol_eq-for-eps(k)-relax} has two solutions in $[0,+\infty)$, as stated in Theorem \ref{th:concentration-VS-relaxation}, translates into that equation \eqref{eq:epsLU_eq} has two solutions in $(-\infty,1]$, namely $\underline{\eps}(k)$ and $\overline{\eps}(k)$. Observing that the left-hand side of \eqref{eq:epsLU_eq} is equal to $\beta/2N > 0$ for $v=1$, while the right-hand side is zero at the same point, we then conclude that, when running backward from $1$ to $-\infty$, the left-hand side is first above, then below, and then above again of the right-hand side,
\begin{figure}[h!]
	\centering
	\includegraphics[width=0.95\columnwidth]{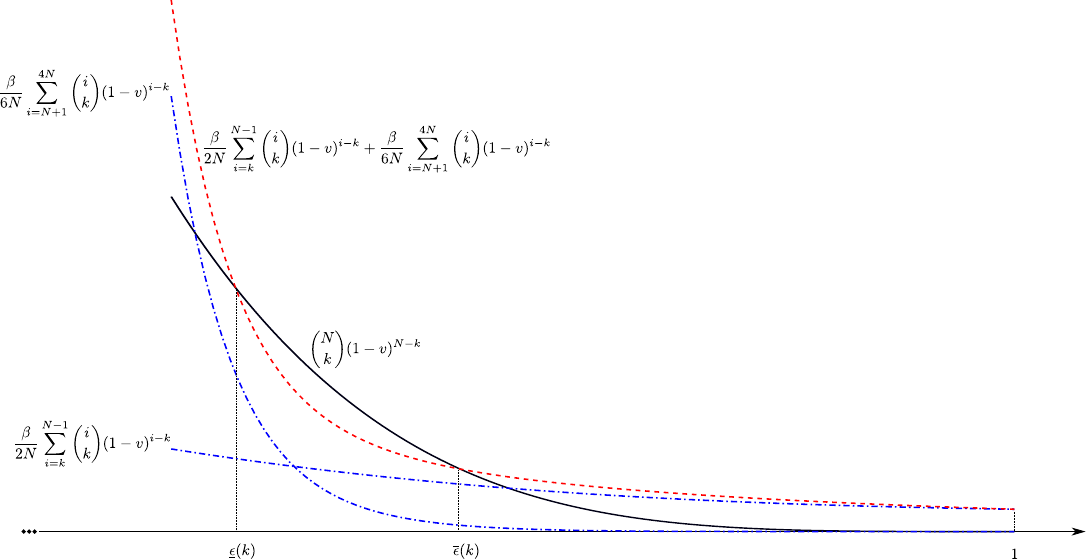}
	\caption{Graphical representation of functions in proof of Theorem \ref{th:explicit}.}
	\label{fig:eqn_epsLU}
\end{figure}
as graphically illustrated in Figure \ref{fig:eqn_epsLU}. \\
Next consider the following two inequalities
\begin{eqnarray}
\frac{\beta}{2N} \sum_{i=k}^{N-1} {i \choose k}(1-v)^{i-k} & \geq & {N \choose k}(1-v)^{N-k}, \label{eq:upper_eq} \\
\frac{\beta}{6N} \sum_{i=N+1}^{4N} {i \choose k}(1-v)^{i-k} & \geq & {N \choose k}(1-v)^{N-k}. \label{eq:lower_eq}
\end{eqnarray}
These two inequalities can be used to effectively locate a suitable upper-bound for $\overline{\eps}(k)$ (inequality \eqref{eq:upper_eq}) and lower-bound for $\underline{\eps}(k)$ (inequality \eqref{eq:lower_eq}). This is explained as follows. \\
Take the ratio of the left-hand side over the right-hand side of equation \eqref{eq:upper_eq}:
$$
\frac{\beta}{2N} \sum_{i=k}^{N-1} \frac{{i \choose k}}{{N \choose k}}(1-v)^{i-N}.
$$
Over $(-\infty,1)$, this function is strictly increasing, moreover for $v=0$ it is smaller than $\beta/2 < 1$ (note that $\frac{{i \choose k}}{{N \choose k}} < 1$) while it tends to $+\infty$ as $v \to 1$. Therefore, it picks the value $1$ in one and only one point in $(0,1)$, which shows that equality is attained in \eqref{eq:upper_eq} for only one value of $v \in (0,1)$. Hence, the two functions showing up in the left-hand and right-hand sides of \eqref{eq:upper_eq} are mutually positioned as shown in Figure \ref{fig:eqn_epsLU}. \\
Further, it is claimed that any $v$ satisfying \eqref{eq:upper_eq} is an upper-bound to $\overline{\eps}(k)$. Indeed, when moving from equation \eqref{eq:epsLU_eq} to \eqref{eq:upper_eq} we have removed from the left-hand side of \eqref{eq:epsLU_eq} a positive term, so shifting to the right the point where equality is achieved in \eqref{eq:upper_eq}; then, owing to the mutual position of the two functions in \eqref{eq:upper_eq} one immediately sees the correctness of the claim. \\
The inequality condition \eqref{eq:lower_eq} can be studied in full analogy to \eqref{eq:upper_eq} with the only advisory that the role of interval $(0,1)$ is played by $(-\infty,1)$ when considering the second inequality \eqref{eq:lower_eq}. \\

\noindent
\emph{Preliminary calculations} \\
\\
To study \eqref{eq:upper_eq} and \eqref{eq:lower_eq}, we shall use a re-writing of the right-hand sides of these inequalities as given in the following. \\
Let
$$
\varphi_{H,k}(v) = \sum_{i=k}^{H-1} {i \choose k} (1-v)^{i-k}.
$$
Notice first that, for $k = 0$, we have
$\varphi_{H,0}(v) = \sum_{i=0}^{H-1} (1-v)^i = \frac{1-(1-v)^H}{v}$.  Next, for $k \leq H-1$, a direct verification proves the validity of the following updating rule
\begin{equation}
\label{update-phi}
\varphi_{H,k}(v) = -\frac{1}{k} \frac{\dd}{\dd v} \varphi_{H,k-1}(v).
\end{equation}
A repeated use (a cumbersome but straightforward exercise) of \eqref{update-phi} now gives
\begin{eqnarray}
\varphi_{H,k}(v) & = & \frac{1-\sum_{i=0}^k {H \choose i}v^i(1-v)^{H-i}}{v^{k+1}} \label{eq:phi_via_Beta-in} \\
& = & \frac{\sum_{i=k+1}^H {H \choose i}v^i(1-v)^{H-i}}{v^{k+1}}. \label{eq:phi_via_Beta}
\end{eqnarray}

\noindent
\emph{Upper bounding $\overline{\eps}(k)$} \\
\\
Substituting \eqref{eq:phi_via_Beta-in} in \eqref{eq:upper_eq}, \eqref{eq:upper_eq} becomes
\begin{equation} \label{eq:upper_eq_new}
\frac{\beta}{2} \left(1-\sum_{i=0}^k {N \choose i}v^i(1-v)^{N-i}\right) \geq N{N \choose k} v^{k+1}(1-v)^{N-k}.
\end{equation}
If we further decrease the left-hand side (and increase the right-hand side) we obtain an inequality the solutions of which are still upper-bounds to $\overline{\eps}(k)$. Starting with the left-hand side, we apply an argument first used in \cite{Alamoetal2015} and, for any $a>1$, write:
\begin{eqnarray}
\lefteqn{\sum_{i=0}^k {N \choose i}v^i(1-v)^{N-i}} \nonumber \\
& \leq & a^k \sum_{i=0}^k {N \choose i}\left( \frac{v}{a} \right)^i(1-v)^{N-i} \nonumber \\
& \leq & a^k \sum_{i=0}^N {N \choose i}\left( \frac{v}{a} \right)^i(1-v)^{N-i} \nonumber \\
& = & a^k \left(1-v+\frac{v}{a}\right)^N \nonumber \\
& = & (1-(1-a))^k \left(1-\frac{a-1}{a}v\right)^N \nonumber \\
& \leq & \ex^{-(1-a)k} \ex^{-\frac{a-1}{a}vN}, \label{bound-left}
\end{eqnarray}
where the last inequality follows from relation $1-z \leq \ex^{-z}$. Similarly, using also the fact that $N{N \choose k} \leq (k+1){N+1 \choose k+1}$,
\begin{eqnarray}
\lefteqn{N{N \choose k} v^{k+1}(1-v)^{N-k}} \nonumber \\
& \leq & (k+1){N+1 \choose k+1} v^{k+1}(1-v)^{N+1-(k+1)} \nonumber \\
& \leq & (k+1) \sum_{i=0}^{k+1} {N+1 \choose i} v^i(1-v)^{N+1-i} \nonumber \\
& \leq & (k+1) \ex^{-(1-a)(k+1)} \ex^{-\frac{a-1}{a}v(N+1)} \nonumber \\
& \leq & (k+1) \ex^{-(1-a)} \ex^{-(1-a)k} \ex^{-\frac{a-1}{a}vN}. \label{bound-right}
\end{eqnarray}
Suppose now $k > 0$ (the case $k = 0$ will be considered separately) and take $a = 1 + 1/\sqrt{k}$. Using \eqref{bound-left} and \eqref{bound-right} in \eqref{eq:upper_eq_new} yields that any $v$ coming from the inequality
$$
\frac{\beta}{2} \left(1-\ex^{\sqrt{k}} \ex^{-\frac{vN}{\sqrt{k}+1}} \right) \geq (k+1) \ex^{\frac{1}{\sqrt{k}}} \ex^{\sqrt{k}} \ex^{- \frac{vN}{\sqrt{k}+1}}
$$
is an upper bound to $\overline{\eps}(k)$. This inequality is equivalent to
$$
\frac{\beta}{2(k+1)} \geq \ex^{\sqrt{k}} \ex^{- \frac{vN}{\sqrt{k}+1}} \left[ \frac{\beta}{2(k+1)} + \ex^{\frac{1}{\sqrt{k}}} \right]
$$
and, solving for $v$, we obtain
$$
v \geq \frac{k}{N} + \frac{\sqrt{k}+1}{N} \left( \lambda + \ln \frac{2}{\beta} + \ln(k+1) \right),
$$
where $\lambda = \ln\left[ \frac{\beta}{2(k+1)} + \ex^{\frac{1}{\sqrt{k}}} \right] + \frac{\sqrt{k}}{\sqrt{k}+1}$. This shows that
$$
\overline{\eps}(k) \leq \frac{k}{N} + \frac{\sqrt{k}+1}{N} \left( \lambda + \ln \frac{2}{\beta} + \ln(k+1) \right)
$$
and the validity of \eqref{eq:epsU<=} (for $k \neq 0,N$ -- recall that we started from equation \eqref{pol_eq-for-eps(k)-relax} that holds for $k < N$ and further left behind the case k=0) follows by noticing that $\lambda \leq 2$. \\
Turn now to the remaining cases, $k = 0$ or $k=N$. \\
Case $k = N$ is trivial because $\overline{\eps}(N) = 1$, which is clearly in agreement with \eqref{eq:epsU<=}. \\
As for $k = 0$, go back to \eqref{eq:upper_eq_new} and use in it \eqref{bound-left} and \eqref{bound-right} with $a = 1 + 1/\sqrt{k + 1}$, which, after substituting $k = 0$, gives $a = 2$ (adding $1$ to $k$ serves the purpose of avoiding division by zero). Operating the same manipulations as before we now obtain
$$
v \geq \frac{2}{N} \left(\ln\left[ \frac{\beta}{2} + \ex \right] + \ln \frac{2}{\beta} \right),
$$
which has the form of the upper bound for $\overline{\eps}(k)$ given in Theorem \ref{th:explicit}. \\
\\
\noindent
\emph{Lower bounding $\underline{\eps}(k)$} \\
\\
First, we want to claim that for any $k$ large enough there is a positive $v$ satisfying equation \eqref{eq:lower_eq}. In fact, for $v = 0$ equation \eqref{eq:lower_eq} reduces to $\frac{\beta}{6N} \sum_{i=N+1}^{4N} {i \choose k} \geq{N \choose k}$ and, using the hockey-stick identity (i.e., $\sum_{i=r}^n {i \choose r} = {n+1 \choose r+1}$), we have
\begin{eqnarray*}
	\lefteqn{\frac{\beta}{6N} \frac{\sum_{i=N+1}^{4N} {i \choose k}}{{N \choose k}}} \\
	& = & \frac{\beta}{6N} \frac{ {4N+1 \choose k+1} - {N+1 \choose k+1} } { {N \choose k}} \\
	& = &  \frac{\beta}{6N} \frac{ (4N+1) \cdots (4N-k+1) - (N+1) \cdots (N-k+1) } {(N) \cdots (N-k+1) \cdot (k+1)} \\
	& \geq &  \frac{\beta}{6} \frac{ 2^{k+1}(N+1) \cdots (N-k+1) - (N+1) \cdots (N-k+1) } {(N+1) \cdots (N-k+1) \cdot (k+1)} \\
	& = & \frac{\beta}{6} \frac{2^{k+1}-1}{k+1},
\end{eqnarray*}
which is greater than $1$ for any
\begin{equation}
\label{k-geq}
k \geq c_1 + c_2 \ln (1/\beta),
\end{equation}
where $c_1$ and $c_2$ are suitable constants. In what follows, we assume that this latter condition is satisfied and hence seek a positive solution of equation \eqref{eq:lower_eq}. \\
The summation in the left-hand side of \eqref{eq:lower_eq} can be re-written as $\sum_{i=N+1}^{4N} {i \choose k}(1-v)^{i-k} = \varphi_{4N+1,k}(v)-\varphi_{N+1,k}(v)$, which, owing to \eqref{eq:phi_via_Beta}, allows us to re-write equation \eqref{eq:lower_eq} as follows
\begin{align} \label{eq:lower_eq_new}
& \frac{\beta}{6} \left(\sum_{i=k+1}^{4N+1} {4N+1 \choose i}v^i(1-v)^{4N+1-i}
-\sum_{i=k+1}^{N+1} {N+1 \choose i}v^i(1-v)^{N+1-i}\right) \nonumber \\
& \quad \quad \geq N{N \choose k} v^{k+1}(1-v)^{N-k},
\end{align}
where moving term $v^{k+1}$ to the right-hand side does not change the inequality sign because $v$ is positive. Similarly to what we did to find an upper bound for $\overline{\eps}(k)$, here we can decrease the left-hand side and increase the right-hand side of \eqref{eq:lower_eq_new} to find a valid lower bound for $\underline{\eps}(k)$. \\
Notice first that $\sum_{i=k+1}^{4N+1} {4N+1 \choose i}v^i(1-v)^{4N+1-i} \geq \frac{1}{2}$ for $v \geq \frac{k+1}{4N+2}$.\footnote{This follows from the fact that $\sum_{i=k+1}^{4N+1} {4N+1 \choose i}v^i(1-v)^{4N+1-i}$ is the cumulative distribution function of a Beta distribution and $\frac{k+1}{4N+2}$ is its mean, which is greater than the median, \cite{PaytonYoungYoung1989}.} Thus, using again the fact $N {N \choose k} \leq (k+1){N+1 \choose k+1}$, we can take
\begin{equation}
\label{eq:lower_eq_new_2}
\frac{\beta}{6} \left(\frac{1}{2}-\sum_{i=k+1}^{N+1} {N+1 \choose i}v^i(1-v)^{N+1-i}\right)
\geq (k+1){N+1 \choose k+1} v^{k+1}(1-v)^{N+1-(k+1)}
\end{equation}
in place of \eqref{eq:lower_eq_new} to obtain a lower bound to $\underline{\eps}(k)$ as long as we impose the additional condition that
\begin{equation}
\label{v-geq}
v \geq \frac{k+1}{4N+2}.
\end{equation}
For any $a > 1$, we now have
\begin{eqnarray*}
	\lefteqn{{N+1 \choose k+1} v^{k+1}(1-v)^{N+1-(k+1)} } \\
	& \leq & \sum_{i=k+1}^{N+1} {N+1 \choose i}v^i(1-v)^{N+1-i} \\
	& \leq & \frac{1}{a^k} \sum_{i=k+1}^{N+1} {N+1 \choose i} (a v)^i(1-v)^{N+1-i} \\
	& \leq & \frac{1}{a^k} \sum_{i=0}^{N+1} {N+1 \choose i} (a v)^i(1-v)^{N+1-i} \\
	& = & \frac{1}{a^k} \left(1+(a-1)v \right)^{N+1} \\
	& \leq & \frac{\ex^{(a-1)v(N+1)}}{a^k},
\end{eqnarray*}
where the last inequality follows from relation $1+z \leq \ex^{z}$. Assume $k > 0$ and take $a = 1+1/\sqrt{k}$. Using the above chain of inequalities twice in \eqref{eq:lower_eq_new_2} (for the term in the left-hand side of \eqref{eq:lower_eq_new_2} we use the inequality obtained by comparing the second with the last term in the chain), we obtain the following condition that is more restrictive than \eqref{eq:lower_eq_new_2}
$$
\frac{\beta}{6} \left(\frac{1}{2}-\frac{\ex^{\frac{v(N+1)}{\sqrt{k}}}}{(1+\frac{1}{\sqrt{k}})^k} \right) \geq (k+1) \frac{\ex^{\frac{v(N+1)}{\sqrt{k}}}}{(1+\frac{1}{\sqrt{k}})^k}.
$$
This inequality is equivalent to
$$
\frac{\beta}{12(\frac{\beta}{6}+k+1)} \geq \frac{\ex^{\frac{v(N+1)}{\sqrt{k}}}}{(1+\frac{1}{\sqrt{k}})^k},
$$
which, solved for $v$, gives
$$
v \leq \frac{k}{N+1} \ln\left[\left(1+\frac{1}{\sqrt{k}}\right)^{\sqrt{k}}\right]
- \frac{\sqrt{k}}{N+1} \left( \ln \frac{12}{\beta} + \ln \left( \frac{\beta}{6}+k+1 \right) \right).
$$
Noticing now that $\ln(1 + x) \geq x - x^2/2$ for all $x \geq 0$, we can finally replace the latter inequality with
\begin{equation}
v \leq \frac{k}{N+1} \left( 1 - \frac{1}{2 \sqrt{k}} \right)
- \frac{\sqrt{k}}{N+1} \left( \ln \frac{12}{\beta} + \ln \left(\frac{\beta}{6}+k+1 \right) \right),
\label{bound-on-v}
\end{equation}
which, for a more handy use, we also rewrite as
$$
v \leq \frac{k}{N} - g(k,N,\beta),
$$
where function $g(k,N,\beta)$ is just the difference between $k/N$ and the right-hand side of \eqref{bound-on-v}. Notice also that this equation is valid also for $k=N$ since \eqref{pol_eq-for-eps(N)-relax} also leads to \eqref{eq:lower_eq}, which has been our starting point in the derivation.

To conclude the proof, we have to put together all inequalities that limit the choice of $v$, namely:

\begin{itemize}
	\item[(i)] $k \geq c_1 + c_2 \ln (1/\beta)$ \quad (equation \eqref{k-geq});
	\item[(ii)] $v \geq \frac{k+1}{4N+2}$ \quad (equation \eqref{v-geq});
	\item[(iii)] $v \leq \frac{k}{N} - g(k,N,\beta)$.
\end{itemize}

Recall that (iii) makes sense only for $k \neq 0$, however this is of no concern because the case $k = 0$ takes care of itself since Theorem \ref{th:explicit} claims that $\underline{\eps}(0) \geq 0$ which is in agreement with the value of $\underline{\eps}(0)$ given in Theorem \ref{th:concentration-VS-relaxation}. For the time being, leave (i) behind. Now, one can take the value of $v$ that achieves equality in (iii), i.e., $v = \frac{k}{N} - g(k,N,\beta)$, provided that this is compatible with (ii), that is, $\frac{k}{N} - g(k,N,\beta) \geq \frac{k+1}{4N+2}$. This can be re-written as $g(k,N,\beta) \leq \frac{k}{N} - \frac{k+1}{4N+2}$. Instead, for those values of $k,N,\beta$ for which this latter inequality does not hold, we have $g(k,N,\beta) > \frac{k}{N} - \frac{k+1}{4N+2}$, from which an easy calculation shows that $2g(k,N,\beta) \geq \frac{k}{N}$, or, equivalently, $\frac{k}{N} - 2g(k,N,\beta) \leq 0$. Since $\underline{\eps}(k) \geq 0$, we conclude that in any case $\underline{\eps}(k) \geq \frac{k}{N} - 2g(k,N,\beta)$. Noticing now that $g(k,N,\beta)$ can be upper bounded by $C' \frac{\sqrt{k} \ln \frac{1}{\beta} + \sqrt{k} \ln k + 1 }{N}$ for a suitable value of the constant $C'$, we conclude that
\begin{equation}
\label{quasi-final}
\underline{\eps}(k) \geq \frac{k}{N} - C \frac{\sqrt{k} \ln \frac{1}{\beta} + \sqrt{k} \ln k + 1 }{N}, \quad \mbox{ where } C = 2 C'.
\end{equation}
Turn now back to consider (i). Condition (i) is not satisfied when $\frac{k}{N} <  (c_1 + c_2 \ln (1/\beta))/N$. However, this latter condition implies that the right-hand side of \eqref{quasi-final} is negative (possibly after enlarging the constant $C$ in \eqref{quasi-final} to a value that, with a little abuse of notation, we still call $C$), so that \eqref{quasi-final} is always a valid lower bound because $\underline{\eps}(k)$ is always non-negative. This concludes the proof. \qed

\section{Proof that $\tilde{\eps}(c,0) \geq 1-(1-c/N)(c/N)^{\frac{c/N}{1-c/N}}$ Asymptotically} \label{appendix:lower_bound_eps_tilde}

The proof is based on the following bounds to the factorial of an integer, \cite{Robbins1955}:
\begin{equation}\label{Robbins_ineq}
\sqrt{2\pi}n^{n+\frac{1}{2}} e^{-n} e^{\frac{1}{12n+1}} < n! < \sqrt{2\pi}n^{n+\frac{1}{2}} e^{-n} e^{\frac{1}{12n}}.
\end{equation}
Start by noticing that
\begin{eqnarray*}
	\tilde{\eps}(c,0) & = & - \ln\left[ \left( \frac{\beta}{N{N \choose c}} \right)^{\frac{1}{N-c}} \right] \\
	& \geq & 1 - \left( \frac{\beta}{N{N \choose c}} \right)^{\frac{1}{N-c}} \\
	& \geq & 1 - \left( \frac{\beta}{{N \choose c}} \right)^{\frac{1}{N-c}} \\
	& = & 1 - \left( \frac{\beta c! (N-c)!}{N!} \right)^{\frac{1}{N-c}},
\end{eqnarray*}
where the first inequality holds because $-\ln(x) \geq 1-x$. Using \eqref{Robbins_ineq} to bound the factorials at the numerator and denominator in the last expression now yields
\begin{eqnarray*}
	\tilde{\eps}(c,0) & \geq & 1 - \left( \frac{\beta\sqrt{2\pi}c^{c+\frac{1}{2}} e^{-c} e^{\frac{1}{12c}}\sqrt{2\pi}(N-c)^{N-c+\frac{1}{2}} e^{-N+c} e^{\frac{1}{12(N-c)}}}{\sqrt{2\pi}N^{N+\frac{1}{2}} e^{-N} e^{\frac{1}{12N+1}}} \right)^{\frac{1}{N-c}} \\
	& = & 1 - \left( \beta \sqrt{2\pi} \right)^{\frac{1}{N-c}} \times e^{\left(\frac{1}{12c}+\frac{1}{12(N-c)}-\frac{1}{12N+1}\right)\frac{1}{N-c}} \times \left( \frac{c(N-c)}{N} \right)^{\frac{1}{2(N-c)}} \times \\
	& & \hspace{3cm} \times \left( \frac{c^c(N-c)^{N-c}}{N^N} \right)^{\frac{1}{N-c}} \\
	& = & 1 - \left( \beta \sqrt{2\pi} \right)^{\frac{1}{N-c}} \times e^{\left(\frac{1}{12c}+\frac{1}{12(N-c)}-\frac{1}{12N+1}\right)\frac{1}{N-c}} \times \left( \frac{c(N-c)}{N} \right)^{\frac{1}{2(N-c)}} \times \\ 
	& & \hspace{3cm} \times \left(1-\frac{c}{N}\right)\left( \frac{c}{N} \right)^{\frac{\frac{c}{N}}{1-\frac{c}{N}}}.
\end{eqnarray*}
Take now $c = \mu N$. The first three terms in the product in the last expression tend to $1$ as $N \to \infty$. Whence,
$$
\tilde{\eps}(d,0) \geq  1 - \left(1-\frac{c}{N}\right)\left( \frac{c}{N} \right)^{\frac{\frac{c}{N}}{1-\frac{c}{N}}}
$$
as $N \to \infty$. This concludes the proof. \qed

\section{Proof of Theorem \ref{theorem-violation-SVM}}
\label{sec:proof_thm_vio_SVM}

For analysis purposes, introduce the augmented probability space $(\mathcal{U} \times \{-1,1\}) \times [0,1]$ endowed with the probability $\probQ = \prob \times \probU$, where $\probU$ is the uniform probability on $[0,1]$ that describes the ``heating variable'' $z$. Next, fix a real parameter value $\alpha$ chosen from the countable set $\{1/j\}$, where $j$ is any positive integer, and consider an independent heated data set $\{\mathbf{u}_i, y_i, (1 - \alpha z_i)\}_{i=1}^N$ generated from $((\mathcal{U} \times \{-1,1\}) \times [0,1])^N$. Note that this situation traces back to the actual data generation mechanism when $\alpha \to 0$ because variable $z$ loses its heating role and augmenting $(\mathcal{U} \times \{-1,1\})$ with $[0,1]$ has no effect.

Suppose we run program \eqref{svm} with the heated data set, that is, we run
\begin{align} \label{svm_heated}
\min_{w \in \mathcal{U}, b \in \mathbb{R} \atop \xi_i \geq 0, i=1,\ldots,N} & \quad  \| w \|^2 + \rho \sum_{i=1}^N \xi_i \\
\textrm{\rm subject to:} & \quad (1 - \alpha z_i) - y_i (\langle w,u_i \rangle - b) \leq \xi_i, \ \ i = 1, \ldots ,N, \nonumber
\end{align}
endowed with the same rule adopted in \eqref{svm} to break the tie in case of non-unique solution. Then, existence and uniqueness are preserved and it is further claimed that the non-accumulation Assumption \ref{asmpt: f concentration} also holds. Indeed, with heated values $y$, the non-accumulation condition writes $\probQ \{(1 - \alpha z) - y (\langle w,\mathbf{u} \rangle - b) = 0 \} = 0$, $\forall (w,b) \in \mathcal{U} \times \mathbb{R}$, a condition that is proven by the following calculation:
\begin{eqnarray} \label{eq:heated_non_degeneracy}
\lefteqn{\probQ \{(1 - \alpha z) - y (\langle w,\mathbf{u} \rangle - b) = 0 \} } \\
& = & \probQ \left\{ z = \frac{1 - y (\langle w,\mathbf{u} \rangle - b)}{\alpha} \right\} \nonumber \\
& = & \probQ \left\{ \probQ \left\{ z = \frac{1 - y (\langle w,\mathbf{u} \rangle - b)}{\alpha} \ \Big| \ \mathbf{u},y \right\} \right\} \nonumber \\
& = & 0. \nonumber
\end{eqnarray}
Hence, the result in Theorem \ref{th:concentration-VS-relaxation} can be applied to the heated situation yielding:
\begin{equation}
\label{result-heated}
\probQ^N \{ \underline{\eps}(s^\ast_\alpha) \leq
V_\alpha(w^\ast_\alpha,b^\ast_\alpha)
\leq \overline{\eps}(s^\ast_\alpha) \}
\geq 1 - \beta,
\end{equation}
where subscript $\alpha$ indicates that the solution has been obtained from the heated program \eqref{svm_heated}, $V_\alpha(w,b) = \probQ \{ (\mathbf{u},y,z): (1 - \alpha z) - y (\langle w,\mathbf{u} \rangle - b) > 0 \}$ and $s^\ast_\alpha$ is the number of $(\mathbf{u}_i,y_i,z_i)$'s for which $(1 - \alpha z_i) - y_i (\langle w^\ast_\alpha,\mathbf{u}_i \rangle - b^\ast_\alpha) \geq 0$.

To re-approach the result \eqref{result-heated} that holds for the heated situation with the initial non-heated problem, let us start by introducing the notation $V_0(w,b) := \probQ \{ (\mathbf{u},y,z): 1 - y (\langle w,\mathbf{u} \rangle - b) > 0 \}$ and note that $V(w,b) := \prob\{ (\mathbf{u},y): 1 - y(\langle w,\mathbf{u} \rangle - b) > 0 \} = V_0(w,b)$. For a given $\alpha > 0$, write
\begin{equation}
\label{V0-Valpha}
V_0(w^\ast,b^\ast) = \left( V_0(w^\ast,b^\ast)  - V_\alpha(w^\ast,b^\ast) \right)
+ \left( V_\alpha(w^\ast,b^\ast) - V_\alpha(w^\ast_\alpha,b^\ast_\alpha) \right)
+ V_\alpha(w^\ast_\alpha,b^\ast_\alpha).
\end{equation}
It is claimed that the first two terms in the right-hand side exhibit the following behaviour:
\begin{itemize}
	\item[(i)] for all realizations of $\{(\mathbf{u}_i,y_i,z_i)\}_{i=1}^N$, it holds that $\lim_{\alpha \to 0} (V_0(w^\ast,b^\ast)  - V_\alpha(w^\ast,b^\ast)) = 0$;
	\item[(ii)] for all realizations of $\{(\mathbf{u}_i,y_i,z_i)\}_{i=1}^N$ such that $w^\ast \neq 0$, it holds that $\lim_{\alpha \to 0} (V_\alpha(w^\ast,b^\ast) - V_\alpha(w^\ast_\alpha,b^\ast_\alpha)) = 0$. \\
\end{itemize}

\noindent
\emph{Proof of (i):} Note that $w^\ast$ and $b^\ast$ only depend on the training sequence and are treated as deterministic in the calculations that follow to compute the risks. Let $B_\alpha := \{ (\mathbf{u},y,z): (1 - \alpha z) - y (\langle w^\ast,\mathbf{u} \rangle - b^\ast) > 0 \}$ and $B_0 := \{ (\mathbf{u},y,z): 1 - y (\langle w^\ast,\mathbf{u} \rangle - b^\ast) > 0 \}$. By a direct inspection one can show that $B_{\alpha_1} \subseteq B_{\alpha_2}$ for $\alpha_2 \leq \alpha_1$ and that $B_0 = \cup_\alpha B_\alpha$. Hence, by $\sigma$-additivity, $V_0(w^\ast,b^\ast) = \probQ \{ B_0 \} = \lim_{\alpha \to 0} \probQ \{ B_\alpha \} = \lim_{\alpha \to 0} V_\alpha(w^\ast,b^\ast)$, and claim (i) remains proven. \\
\\
\emph{Proof of (ii):} Note that $w^\ast_\alpha \to w^\ast$ and that $b^\ast_\alpha \to b^\ast$ as $\alpha \to 0$. Moreover, by assumption $w^\ast \neq 0$. Let $B^\alpha_\alpha := \{ (\mathbf{u},y,z): \; (1 - \alpha z) - y (\langle w^\ast_\alpha,\mathbf{u} \rangle - b^\ast_\alpha) > 0 \}$. Over the complement of set $A:= \{(\mathbf{u},y,z): \; 1 - y (\langle w^\ast,\mathbf{u} \rangle - b^\ast) = 0\}$, for any given $(\mathbf{u},y,z)$, the two left-hand sides in the inequalities that define $B_\alpha$ and $B^\alpha_\alpha$ agree in sign in the limit when $\alpha \to 0$, so that, in the limit,  $B_\alpha \triangle B^\alpha_\alpha \subseteq A$ ($\triangle$ denotes symmetric difference). More formally, this means that for all $(\mathbf{u},y,z) \in A^c$, the complement of $A$, there exists an $\bar{\alpha}$ such that $(\mathbf{u},y,z) \notin B_\alpha \triangle B^\alpha_\alpha$ for all $\alpha \leq \bar{\alpha}$. This property in turn implies that $\limsup_{\alpha \to 0} \probQ \{ B_\alpha \triangle B^\alpha_\alpha \} \leq \probQ \{ A \}$ and therefore we have:
\begin{eqnarray*}
	\lefteqn{\limsup_{\alpha \to 0} | V_\alpha(w^\ast,b^\ast) - V_\alpha(w^\ast_\alpha,b^\ast_\alpha) | } \\
	& = & \limsup_{\alpha \to 0} | \probQ \{ B_\alpha \} - \probQ \{ B^\alpha_\alpha \} | \\
	& \leq & \limsup_{\alpha \to 0} \probQ \{ B_\alpha \triangle B^\alpha_\alpha \} \\
	& \leq & 	\probQ \{ A \} \\
	& = & 0 \quad (\mbox{recall that } w^\ast \neq 0 \mbox{ and use Assumption \ref{non-degeneracy-SVM}}).
\end{eqnarray*}
This completes the proof of (ii).\footnote{Note that Assumption \ref{non-degeneracy-SVM} cannot be dispensed for as shown by the following counterexample. Suppose that $u \in \mathbb{R}$ has mass concentrated over $\pm 1$ with equal probability $0.5$ and $y=u$. Clearly, Assumption \ref{non-degeneracy-SVM} is not satisfied in this case. When the $u_i$ are not picked all equal and $\rho$ is large, we have $w^\ast = 1$ and $b^\ast = 0$, and $V_\alpha(w^\ast,b^\ast)$ is zero. However, with the exception of a zero-probability set, we have $w^\ast_\alpha \cdot 1 - b^\ast_\alpha < 1$ and $w^\ast_\alpha \cdot (-1) - b^\ast_\alpha > -1$, so that $V_\alpha(w^\ast_\alpha,b^\ast_\alpha) \neq 0$ with a value that depends on the realization of  $\{(\mathbf{u}_i,y_i,z_i)\}_{i=1}^N$, but that is constant with $\alpha$. Hence, $\lim_{\alpha \to 0} (V_\alpha(w^\ast,b^\ast) - V_\alpha(w^\ast_\alpha,b^\ast_\alpha)) \neq 0$. } \\
\\
Using (i) and (ii) in \eqref{V0-Valpha}, we obtain:
\begin{equation} \label{convergence-V-w-not-0}
\left.
\begin{array}{c}
\parbox{0.8\columnwidth}
{for all realizations of $\{(\mathbf{u}_i,y_i,z_i)\}_{i=1}^N$ such that $w^\ast \neq 0$, it holds that} \\
\lim_{\alpha \to 0} V_\alpha(w^\ast_\alpha,b^\ast_\alpha) = V_0(w^\ast,b^\ast).
\end{array}
\right\}
\end{equation}
Turn now to consider $s^\ast$ and $s^\ast_\alpha$. We show that:
\begin{equation} \label{convergence-s-w-not-0}
\left.
\begin{array}{c}
\parbox{0.8\columnwidth}{with the exception of a zero-probability set, for all realizations of $\{(\mathbf{u}_i,y_i,z_i)\}_{i=1}^N$ such that $w^\ast \neq 0$, it holds that}\\
\lim_{\alpha \to 0 } s^\ast_\alpha = s^\ast.
\end{array}
\right\}
\end{equation}
To see this, note that, when $w^\ast \neq 0$ and with the exception of a zero-probability set, Assumption \ref{non-degeneracy-SVM} implies that the $(\mathbf{u}_i,y_i,z_i)$ such that $1 - y_i (\langle w^\ast,\mathbf{u}_i \rangle - b^\ast) \geq 0$ correspond to the active constraints for \eqref{svm}, and all of these active constraints are strictly needed to determine the solution $w^\ast,b^\ast,\xi_i^\ast$. A small enough heating keeps these and only these constraints active for \eqref{svm_heated} too, which implies that $s^\ast_\alpha = s^\ast$ for all $\alpha$ small enough.

Using \eqref{result-heated}, \eqref{convergence-V-w-not-0}, and \eqref{convergence-s-w-not-0}, we are now ready to establish results that quantify the violation when $w^\ast \neq 0$.

Let $I(k) = [\underline{\eps}(k),\overline{\eps}(k)]$ and define the following events in $((\mathcal{U} \times \{-1,1\}) \times [0,1])^N$:
\begin{eqnarray*}
	E & = &  \big\{ \{ (\mathbf{u}_i,y_i,z_i)\}_{i=1}^N: \; w^\ast \neq 0 \; \wedge V(w^\ast,b^\ast) \notin I(s^\ast) \big\} \\
	E_\alpha & = & \big\{ \{ (\mathbf{u}_i,y_i,z_i)\}_{i=1}^N: \; w^\ast \neq 0 \; \wedge V_\alpha(w^\ast_\alpha,b^\ast_\alpha) \notin I(s^\ast_\alpha) \big\} \\
	E_\alpha^+ & = & \cap_{\alpha' \leq \alpha} E_{\alpha'}.
\end{eqnarray*}
Using \eqref{convergence-V-w-not-0} and \eqref{convergence-s-w-not-0}, one can easily show that
$$
E \subseteq \bigcup_\alpha E_\alpha^+,
$$
from which we obtain
\begin{eqnarray*}
	\lefteqn{\prob^N \{ w^\ast \neq 0 \; \wedge \; V(w^\ast,b^\ast) \notin I(s^\ast) \} } \\
	& = & \probQ^N (E) \\
	& \leq &  \probQ^N (\cup_\alpha E_\alpha^+) \\
	& = & \lim_{\alpha \to 0} \probQ^N( E_\alpha^+ ) \quad (\mbox{since } E_\alpha^+ \mbox{ is increasing as } \alpha \mbox{ decreases})\\
	& \leq & \limsup_{\alpha \to 0} \probQ^N( E_\alpha ) \quad (\mbox{since } E_\alpha^+ \subseteq E_\alpha)\\
	& = & \limsup_{\alpha \to 0} \probQ^N( w^\ast \neq 0 \; \wedge \; V_\alpha(w^\ast_\alpha,b^\ast_\alpha) \notin I(s^\ast_\alpha) ) \\
	& \leq & \limsup_{\alpha \to 0} \probQ^N( V_\alpha(w^\ast_\alpha,b^\ast_\alpha) \notin I(s^\ast_\alpha) ).
\end{eqnarray*}
Applying \eqref{result-heated} to the last term finally gives
\begin{equation} \label{eq:vio_prob_w_not_0}
\prob^N \{ w^\ast \neq 0 \; \wedge \; V(w^\ast,b^\ast) \notin I(s^\ast) \} \leq \beta.
\end{equation}

To conclude the proof, we have now to account for the realizations of $\{(\mathbf{u}_i,y_i,z_i)\}_{i=1}^N$ for which $w^\ast = 0$ and show that
\begin{equation} \label{eq:vio_prob_w=0}
\prob^N \{ w^\ast = 0 \; \wedge \; V(w^\ast,b^\ast) \notin I(s^\ast) \} \leq 2 \beta.
\end{equation}
In fact, \eqref{eq:vio_prob_w_not_0} and \eqref{eq:vio_prob_w=0} together give
\begin{eqnarray*}
	\lefteqn{\prob^N \{  V(w^\ast,b^\ast) \notin I(s^\ast) \}} \\
	& = & \prob^N \{ w^\ast \neq 0 \; \wedge \; V(w^\ast,b^\ast) \notin I(s^\ast) \} + \; \prob^N \{ w^\ast = 0 \; \wedge \; V(w^\ast,b^\ast) \notin I(s^\ast) \} \\
	& \leq & 3 \beta,
\end{eqnarray*}
which is equivalent to the statement of Theorem \ref{theorem-violation-SVM}.

To prove \eqref{eq:vio_prob_w=0}, first notice that substituting $w^\ast = 0$ in program \eqref{svm} gives
\begin{align*}
\min_{b \in \mathbb{R} \atop \xi_i \geq 0, i=1,\ldots,N} & \quad  \rho \sum_{i=1}^N \xi_i \\
\textrm{\rm subject to:} & \quad 1 + y_i b \leq \xi_i, \ \ i = 1, \ldots ,N, \nonumber
\end{align*}
and a simple direct inspection reveals that at optimum either $b^\ast = -1$ (when no. of $y_i=1$ $\geq$ no. of $y_i=-1$; notice that when these two numbers are equal, $b^\ast = -1$ is enforced by the adopted tie-break rule) or $b^\ast = 1$ (when no. of $y_i=1$ $<$ no. of $y_i=-1$). The analysis is thus split into two sub-cases, namely, $(w^\ast = 0,b^\ast = -1)$ and $(w^\ast = 0,b^\ast = 1)$, and \eqref{eq:vio_prob_w=0} is obtained by showing that
$$
\prob^N \{ w^\ast = 0 \; \wedge \; b^\ast = \odot \; \wedge \; V(w^\ast,b^\ast) \notin I(s^\ast) \} \leq \beta
$$
where $\odot$ is either $-1$ or $1$.

The proof for one case is identical to that for the other. Choose thus one, say $(w^\ast = 0,b^\ast = -1)$, and consider a version of the heated program \eqref{svm_heated} where $w$ and $b$ are always (i.e. for all realizations of $\{(\mathbf{u}_i,y_i,z_i)\}_{i=1}^N$) constrained to take the values $0$ and $-1$, respectively:
\begin{align} \label{svm_heated_w=0_b=-1}
\min_{w = 0, b = -1 \atop \xi_i \geq 0, i=1,\ldots,N} & \quad  \| w \|^2 + \rho \sum_{i=1}^N \xi_i \\
\textrm{\rm subject to:} & \quad (1 - \alpha z_i) - y_i (\langle w,u_i \rangle - b) \leq \xi_i, \nonumber \\
& \quad i = 1, \ldots ,N, \nonumber
\end{align}
which is equivalent to
\begin{align*}
\min_{\xi_i \geq 0, i=1,\ldots,N} & \quad \rho \sum_{i=1}^N \xi_i \\
\textrm{\rm subject to:} & \quad (1 - \alpha z_i) - y_i \leq \xi_i, \ \ i = 1, \ldots ,N. \nonumber
\end{align*}
Program \eqref{svm_heated_w=0_b=-1} is quite a peculiar instance of \eqref{scenario program-relaxed}, since $x = (w,b)$ belongs to a vector space with null dimensionality. Still, the theory of Section \ref{sec:theory} retains its validity. As a matter of fact,  \eqref{svm_heated_w=0_b=-1} has clearly a unique solution, which is
$$
\tilde{w}^\ast_\alpha = 0, \; \tilde{b}^\ast_\alpha = -1, \; \tilde{\xi}^\ast_{i,\alpha} = (1-\alpha z_i)-y_i,
$$
and it satisfies the non-accumulation Assumption \ref{asmpt: f concentration} (as shown by \eqref{eq:heated_non_degeneracy} with $w = 0$ and $b = -1$). Theorem \ref{th:concentration-VS-relaxation} can therefore be applied to \eqref{svm_heated_w=0_b=-1} yielding
\begin{equation} \label{result-heated_w=0_b=-}
\probQ^N \{ V_\alpha(\tilde{w}^\ast_\alpha,\tilde{b}^\ast_\alpha) \notin I(\tilde{s}^\ast_\alpha) \}
\leq \beta,
\end{equation}
for all $\alpha$, where $V_\alpha(\tilde{w}^\ast_\alpha,\tilde{b}^\ast_\alpha) = \probQ \{ (\mathbf{u},y,z): (1 - \alpha z) - y (\langle \tilde{w}^\ast_\alpha,\mathbf{u} \rangle - \tilde{b}^\ast_\alpha) > 0 \} = \probQ \{ (\mathbf{u},y,z): y < (1 - \alpha z) \}$ and $\tilde{s}^\ast_\alpha$ is the number of $(\mathbf{u}_i,y_i,z_i)$ for which $(1 - \alpha z_i) - y_i (\langle \tilde{w}^\ast_\alpha ,\mathbf{u}_i \rangle - \tilde{b}^\ast_\alpha) \geq 0$, i.e. for which $y_i \leq (1 - \alpha z_i)$.

Recalling that $\alpha = 1/j$, with $j$ any positive integer, that $y$ can be either $1$ or $-1$, and that $z \in [0,1]$, one sees that for all the realizations of $\{(\mathbf{u}_i,y_i,z_i)\}_{i=1}^N$ such that $w^\ast = 0$ and $b^\ast = -1$ and for all $\alpha$, it holds  that
\begin{eqnarray*}
	V(w^\ast,b^\ast) & = & V(0,-1) \\
	& = & \prob \{ (\mathbf{u},y): y < 1 \} \\
	& = & \probQ \{ (\mathbf{u},y,z): y < 1 \} \\
	& = & \probQ \{ (\mathbf{u},y,z): y < (1 - \alpha z) \} \\
	& = & V_\alpha(\tilde{w}^\ast_\alpha,\tilde{b}^\ast_\alpha).
\end{eqnarray*}
and, with exception of when $z_i=0$ for some $i$, which has zero-probability, that
\begin{eqnarray*}
	s^\ast & = & \mbox{no. of } y_i = -1 \quad (\mbox{recall how } s^\ast \mbox{ is defined when } w^\ast=0)\\
	& = & \mbox{no. of } y_i \leq (1 - \alpha z_i) \\
	& = & \tilde{s}^\ast_\alpha.
\end{eqnarray*}
Hence, we have
\begin{eqnarray*}
	\lefteqn{\prob^N \{ w^\ast = 0 \; \wedge \; b^\ast = -1 \; \wedge \; V(w^\ast,b^\ast) \notin I(s^\ast) \}} \\
	& = & \probQ^N \{ w^\ast = 0 \; \wedge \; b^\ast = -1 \; \wedge \; V(w^\ast,b^\ast) \notin I(s^\ast) \} \\
	& = & \probQ^N \{ w^\ast = 0 \; \wedge \; b^\ast = -1 \; \wedge \; V_\alpha(\tilde{w}^\ast_\alpha,\tilde{b}^\ast_\alpha) \notin I(\tilde{s}_\alpha^\ast) \} \\
	& \leq & \probQ^N \{ V_\alpha(\tilde{w}^\ast_\alpha,\tilde{b}^\ast_\alpha) \notin I(\tilde{s}_\alpha^\ast) \} \\
	& \leq & \beta,
\end{eqnarray*}
which is the sought relation. The same argument applies \emph{mutatis mutandis} for the case $(w^\ast=0,b^\ast=1)$.

This concludes the proof. \qed

\vskip 0.3in

\bibliography{21-0641}

\end{document}